\documentclass[runningheads]{llncs}
\usepackage{graphicx}

\usepackage{placeins}

\usepackage{amsmath,amssymb} % define this before the line numbering.
\usepackage{bm} % Useful to make bold Greek letters.
\usepackage{amsfonts}
\usepackage{times}
\usepackage{float}
\usepackage{color}
\usepackage{xcolor}

\usepackage{array}
\newcolumntype{C}[1]{>{\centering\let\newline\\\arraybackslash\hspace{0pt}}m{#1}}

\usepackage{xspace}
\makeatletter
\DeclareRobustCommand\onedot{\futurelet\@let@token\@onedot}
\def\@onedot{\ifx\@let@token.\else.\null\fi\xspace}
\def\eg{\emph{e.g}\onedot} 
\def\ie{\emph{i.e}\onedot}

\def\etal{\emph{et al}\onedot}
\makeatother

\makeatletter
\newcommand\boldparagraph{\@startsection{paragraph}{4}{\z@}%
                       {4\p@ \@plus 2\p@ \@minus 2\p@}%
                       {-0.5em \@plus -0.22em \@minus -0.1em}%
                       {\normalfont\normalsize\bfseries\boldmath}}
\makeatother

\usepackage[tight,bf]{subfigure}
\newlength{\figwidth}

\begin{document}
%===========================================================

\title{Gaussian Process Deep Belief Networks:\\A Smooth Generative Model of Shape with Uncertainty Propagation}
\titlerunning{Gaussian Process Deep Belief Networks} % Replace an abstracted version of your paper's title here

%===========================================================

\author{Alessandro Di Martino\,\inst{1} \and %\orcidID{0000-0001-5213-9571}
Erik Bodin\,\inst{2} \and %\orcidID{0000-0002-5835-2004}
Carl Henrik Ek\,\inst{2} \and %\orcidID{0000-0003-1302-6309}
Neill D.\,F. Campbell\,\inst{1}} % \orcidID{0000-0003-2130-4903}
%
%Please include author names in full in the paper,
%If any authors have names that can be parsed into FirstName LastName in multiple ways, please include the correct parsing, in a comment to the volume editors:
%\index{Lastnames, Firstnames}
%\index{Di Martino, Alessandro}
%\index{Bodin, Erik}
%\index{Ek, Carl Henrik}
%\index{Campbell, Neill D. F.}

\authorrunning{A. Di Martino et al.~~--~~Paper Accepted at ACCV2018} % A shorter version of authors' name

%===========================================================

\institute{University of Bath, Department of Computer Science \and
University of Bristol, Department of Computer Science}

\maketitle

%===========================================================
\begin{abstract}
The shape of an object is an important characteristic for many vision problems such as segmentation, detection and tracking. Being independent of appearance, it is possible to generalize to a large range of objects from only small amounts of data. However, shapes represented as silhouette images are challenging to model due to complicated likelihood functions leading to intractable posteriors. In this paper we present a generative model of shapes which provides a low dimensional latent encoding which importantly resides on a smooth manifold with respect to the silhouette images. The proposed model propagates uncertainty in a principled manner allowing it to learn from small amounts of data and providing predictions with associated uncertainty. We provide experiments that show how our proposed model provides favorable quantitative results compared with the state-of-the-art while simultaneously providing a representation that resides on a low-dimensional interpretable manifold.

\end{abstract}
%===========================================================

\section{Introduction}

The space of silhouette images is challenging to work with as it is not smooth in terms of a representation as pixels. A transformation that we would consider semantically smooth might correspond to a drastic change in pixel values. Our goal is to learn a smooth low dimensional representation of silhouette images such that images can be generated in a natural manner. Further, as data is at a premium, we want to learn a fully probabilistic model that allows us to propagate uncertainty throughout the generative process. This will allow us to learn from \emph{smaller amounts of data} and also associate a quantified uncertainty to its predictions. This uncertainty allows the model to be used as a building block in larger models.

The results of our model challenge the current trend in unsupervised learning towards maximum likelihood training of increasingly large parametric models with increasingly large datasets. We demonstrate that by propagating uncertainty throughout the model, our approach outperforms two standard generative deep learning models, a Variational Auto-Encoder (VAE~\cite{Kingma2013}) and a Generative Adversarial Network (InfoGAN~\cite{chen2016infogan}) with comparable architectures and can achieve similar performance with far smaller training datasets.

In our work we revisit a few classic machine learning models with complementary properties. On the one hand, parametric models such as Restricted Boltzmann Machines (RBMs)~\cite{smolensky1986parallel} are particularly interesting as they are stochastic, generative and can be stacked easily into \emph{deeper} models such as deep belief networks (DBNs); these can be trained in a greedy fashion, layer by layer \cite{hinton2006reducing}. RBMs can approximate a probability distribution on visible units. DBNs, in addition, learn deep representations by composing features learned by the lower layers, yielding progressively more abstract and flexible representations at higher layers and often leading to more expressive and efficient models compared to shallow ones \cite{bengio2007scaling}.

However, DBNs suffer from a number of limitations. Firstly, they do not guarantee a smooth representation in the learned latent space. Secondly, the classic contrastive divergence algorithm used for greedy training is slow and can place limitations on architectures. Finally, a DBN does not provide any explicit generative process from a manifold, as the standard way to sample from a DBN is to start from a training example and perform iterations of Gibbs sampling.

The Gaussian Process Latent Variable Model (GPLVM)~\cite{lawrence2005probabilistic} combines a Gaussian process (GP) prior with a likelihood function in order to learn a representation. By specifying a prior that encourages smooth functions a smooth latent representation can be recovered. However, to make inference tractable the likelihood is also chosen to be Gaussian which does not reflect the statistics of natural images. Further, even though the mapping from the latent space is non-linear the posterior is linear in the observed space. This makes the GPLVM unsuitable for modelling images. To circumvent this one can compose hierarchies of GPs~\cite{damianou2013deep}, however, these models are inherently difficult to train.

The characteristics of the DBN and GPLVM can be considered complementary, where the DBN excels the GPLVM fails and vice versa.
Unfortunately, combining the two models into a single one by simply stacking a GPLVM on top of a DBN would not preserve uncertainty propagation. Furthermore, this would pose a challenge to training (while the GPLVM is a non-parametric model trained by optimizing an objective function, a DBN is a parametric model, with non-differentiable Bernoulli units, and is trained with contrastive divergence).
Another important challenge is learning from very little data.
The ability to learn from a small dataset expands the applicability of a model to domains where there is a lack of available data or where collection of data is costly or time-consuming.

In this paper we address these challenges and present the following contributions:
\begin{enumerate}
    \item A model (which we call GPDBN) that combines the properties of a smooth, interpretable manifold for synthesis with a data specific likelihood function (a deep structure) capable of decomposing images into an efficient representation while propagating uncertainty throughout the model in a principled manner.
    \item We train the model end to end using back propagation with the same complexity as a standard feed-forward neural network by minimising a single objective function.
    \item We also show that the model is able to learn from very little data, outperforming current generative deep learning models, as well as scaling linearly to larger datasets by the use of mini-batching.
\end{enumerate}

%===========================================================
\section{Related Work}
\label{related_work}

\newcommand*\rot{\rotatebox{90}}

Modelling of shape is important for many computer vision tasks. It is beyond the scope of this paper to make a complete review of the topic, we refer the reader to the comprehensive work of Taylor~\etal~\cite{statModelsOfShape}. In our work we focus on recent unsupervised statistical models that operate directly on the pixel domain.
Interest in these models was revived by the Shape Boltzmann Machine (SBM) work of Eslami~et~al.~\cite{eslami2013shape} and they have been shown to be useful for a variety of vision applications~\cite{eslami2012generative,deepPartBasedGenerativeShape,Vedaldi2015semanticpartsegmentation}.
These deep models can also be readily extended into the 3D domain, \eg,~by recent work on 3D ShapeNets~\cite{shapenets2015}.
Detailed analysis of the DBN, GPLVM and SBM is provided in \S~\ref{background}.

\boldparagraph{Desirable Properties}
Table~\ref{tab:related_work} highlights the desirable properties of the most closely related previous works.
We have identified four advantageous properties:
(i)~It is well known that pixel silhouettes are not well modelled by a Gaussian likelihood.
(ii)~The utility of an unsupervised shape model is well described by the properties of its latent representation.
Ensuring a smooth manifold opens up a number of applications to data in the pixel domain that previously required custom representations, \eg,~interactive drawing~\cite{sketchPaper}.
(iii)~A fully generative model ensures that there is a well defined space that can be sampled as well as interpreted;
\eg, dynamics models can be defined in such a space to perform tracking~\cite{Rotopp2016,victor2012pwp3dsdf}.
(iv)~Correctly propagating uncertainty is vital to perform data efficient learning, for example when data is scarce or expensive to obtain.

\begin{table}[t]
\centering%
\renewcommand{\rot}[1]{#1}
\renewcommand{\arraystretch}{1.2}
\setlength{\tabcolsep}{7.5pt}
\setlength{\tabcolsep}{6pt}
%\small
\scalebox{0.7}{%
\centering%
\begin{tabular}{lcccc}
% &
%\rot{\shortstack[c]{Non-Gaussian\\Likelihood}} &
%\rot{\shortstack[c]{Explicit Smooth\\Low-Dim Manifold}} &
%\rot{\shortstack[c]{Fully\\Generative}} &
%\rot{\shortstack[c]{Propagates\\Uncertainty}} \\[2pt]
 &
Non-Gaussian Likelihood &
Explicit Smooth Low-Dim Manifold &
Fully Generative &
Propagates Uncertainty \\[2pt]
GPLVM~\cite{lawrence2005probabilistic} &  & \checkmark & \checkmark & \checkmark \\
GPLVMDT~\cite{victor2012pwp3dsdf} &  & \checkmark & \checkmark & \checkmark \\
DBN~\cite{hinton2006reducing} & \checkmark & & & \checkmark \\
SBM~\cite{eslami2013shape} & \checkmark & & & \checkmark  \\
VAE~\cite{Kingma2013} & \checkmark & $\sim$ & \checkmark & \\%$\sim$
InfoGAN~\cite{chen2016infogan} & \checkmark & $\sim$ & \checkmark & \\%$\sim$
ShapeOdds~\cite{ShapeOdds} & \checkmark &  & \checkmark & \checkmark \\[2pt]%$\sim$
\textbf{This work} & \checkmark & \checkmark & \checkmark & \checkmark\\
\end{tabular}
}\\[2pt]
\caption{
Summary of properties of related models.
}
\label{tab:related_work}
\end{table}

\boldparagraph{Auto-Encoders}
The VAE model by Kingma and Welling~\cite{Kingma2013} performs a variational approximation of a generative model with a non-Gaussian likelihood through a feed-forward or Multi-Layer Perceptron (MLP) network.
In addition, it uses MLP networks to encode the variational parameters (in a similar manner to~\cite{lawrence2006local}).
While this model provides a generative mapping, the feed-forward (decoder) network fails to propagate uncertainty from the latent space.
Furthermore, the independent prior on the latent space does not promote a smooth manifold;
any smoothness arises as a by-product of the MLP encoding network.
This characteristic depends on the MLP architecture and is not directly parametrised.
The key limitation of the VAE for our purposes is the lack of uncertainty propagation that results in poor results with limited training data.

% This paragraph can go if we are pushed for space.
The guided, non-parametric autoencoder model of Snoek~et~al.~\cite{nonparametricGuidanceAutoEncoder} appears similar, however, there are a number of important differences.
They use label information (supervision) to guide a latent space learning process for an autoencoder;
this is not a pure unsupervised learning task and we do not have label information available to us.
Furthermore, as with the VAE, uncertainty is not propagated from the latent manifold %(over the unsupervised labels)
to the output space due to the use of the feed-forward network to the output.% (the same situation as the VAE).

\boldparagraph{InfoGAN}
Another prominent generative model in unsupervised learning is the Generative Adversarial Network (GAN)~\cite{goodfellow2014gan}. The model learns an implicit generator distribution using a minimax game between a deep generator network, % \emph{G},
which transforms a noise variable %\emph{z}
to a sample, and a deep discriminator network, % \emph{D},
which is used to classify between samples from the generator distribution and the true data distribution.
One issue common with GAN models is that they do not provide a smooth latent manifold for synthesis nor uncertainty in their estimates (like the VAE).
From the plethora of different variations of GANs models available in the literature we have chosen to include in our comparisons the InfoGAN model~\cite{chen2016infogan}, since it also considers the goal of interpretable latent representations (by maximising the mutual information between a subset of GAN's noise variables and observations).

\boldparagraph{ShapeOdds}
The recent ShapeOdds work of Elhabian and Whitaker~\cite{ShapeOdds} confers state-of-the-art performance and captures many of the desired properties including a generative probabilistic model that propagates uncertainty. The approach taken is quite different to ours as they specify a detailed probabilistic model including a Gaussian Markov Random Field (MRF) with individual Bernoulli random variables for the pixel lattice. In contrast, our model is more flexible, we allow the network to learn the structure from the data directly but ensure that we still maintain uncertainty quantification throughout. We would also argue that the specific form of the low dimensional manifold we generate is desirable with its guaranteed smoothness that makes the latent space readily interpretable.
This provides the tradeoff between the two models. We expect the ShapeOdds model to perform very well at generalisation due to the inclusion of the MRF prior. In contrast, our model will be more data dependent in this respect (weaker prior assumptions on the nature of images), however, it provides a generative space that is highly interpretable and easy to work with.
We identify that a topic for further work would be to combine our smooth priors with the likelihood model of ShapeOdds.

\boldparagraph{GPLVM Representations}
A possible workaround to the problem of non-Gaussian likelihoods is to perform a deterministic transformation to a domain where the data is approximately Gaussian.
This has been successful for domains where, for example, the shape can be represented in a new geometric representation away from pixels, \eg,~parametric curves~\cite{CampbellSIGGRAPH14,victor2011nonlinearshapemanifolds}.
However, this is application dependent and not suitable for arbitrary pixel based silhouettes considered here.
A common approach that retains the pixel grid is to transform it into a level-set problem via the distance transform, \eg,~\cite{victor2012pwp3dsdf}.
This can improve results in some settings, however, the uncertainty is not correctly preserved and therefore not correctly captured in predictions.
We denote this model GPLVMDT in our comparisons.

%===========================================================
\section{Background}
\label{background}

%===========================================================
\subsection{Deep Belief Networks}
\label{dbn}

\begin{figure}[t]
\setlength{\figwidth}{0.3\linewidth}
\centering
\begin{minipage}[b]{\figwidth}\centering
    \subfigure[RBM]{%
      \centering%
      \includegraphics[width=0.8\figwidth]{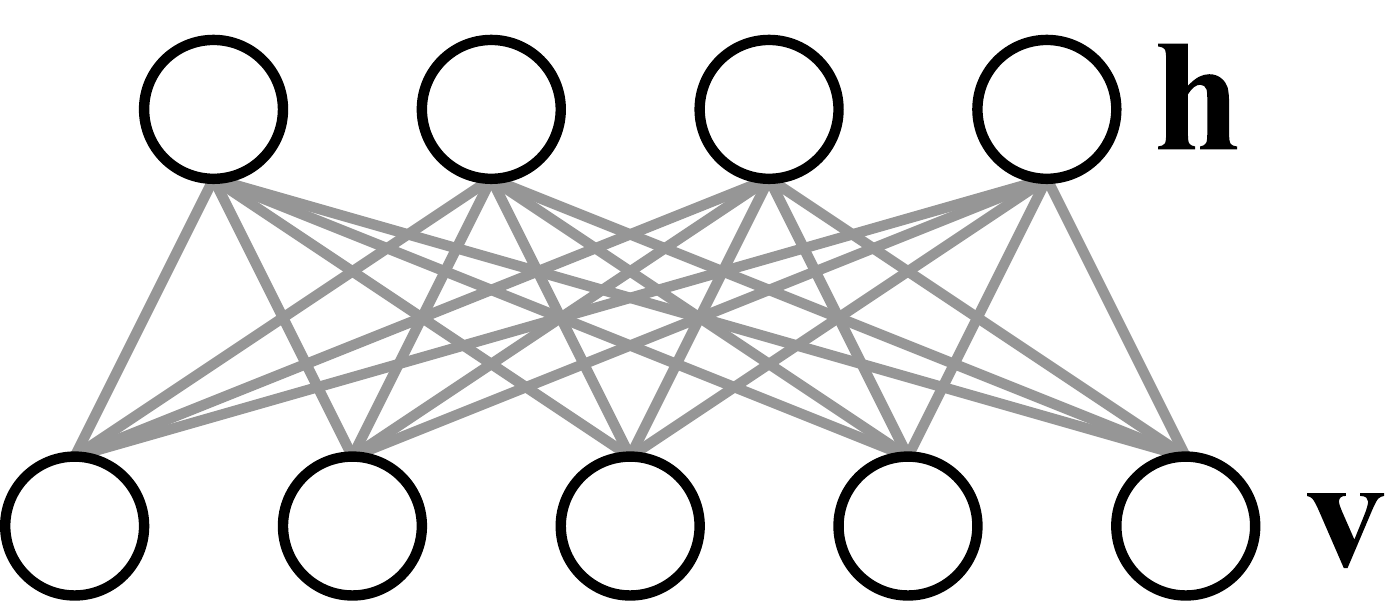}
        \label{fig:rbm}
    }\\%subfigure
    \subfigure[DBN]{%
      \centering%
      \includegraphics[width=0.8\figwidth]{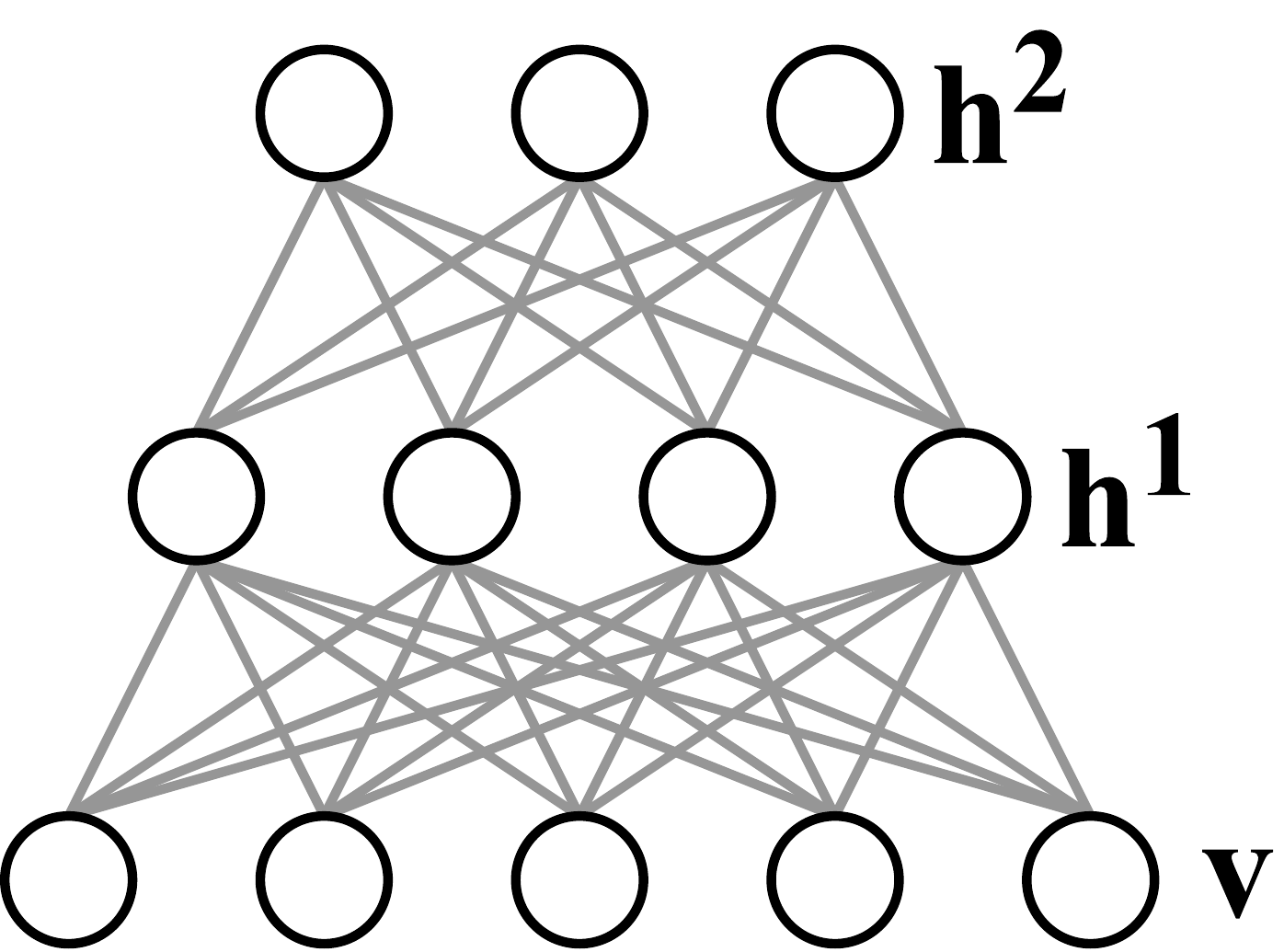}
        \label{fig:dbn}
    }%subfigure
\end{minipage}\hfill%
    \subfigure[SBM]{%
      \centering%
      \includegraphics[width=0.8\figwidth]{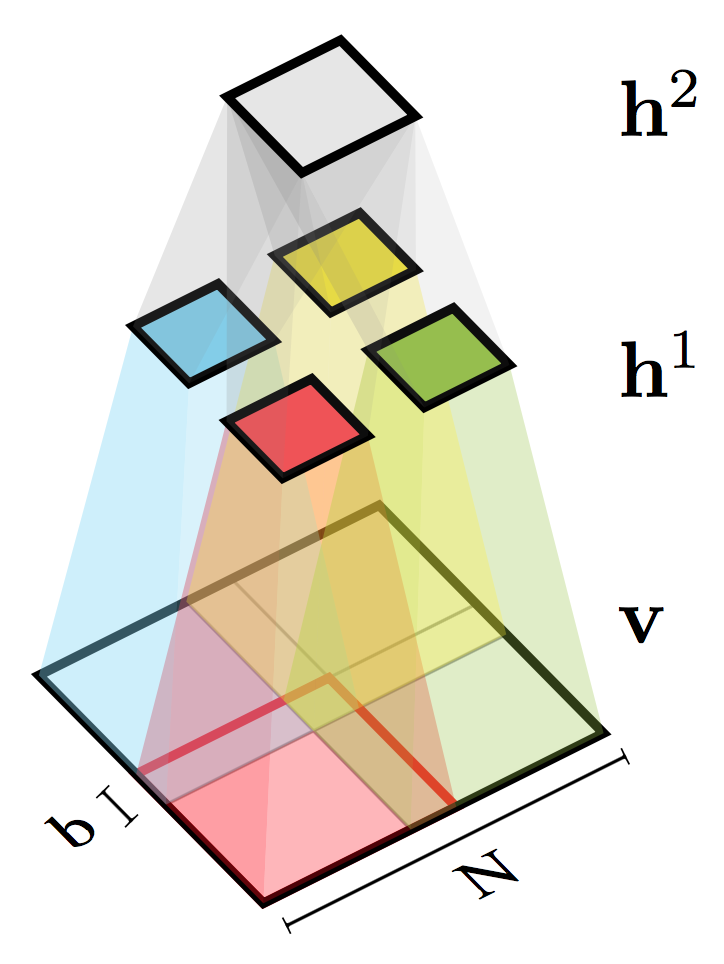}
        \label{fig:sbm}
    }\hfill%subfigure
    \subfigure[Our GPDBN]{%
      \centering%
      \includegraphics[width=1.15\figwidth]{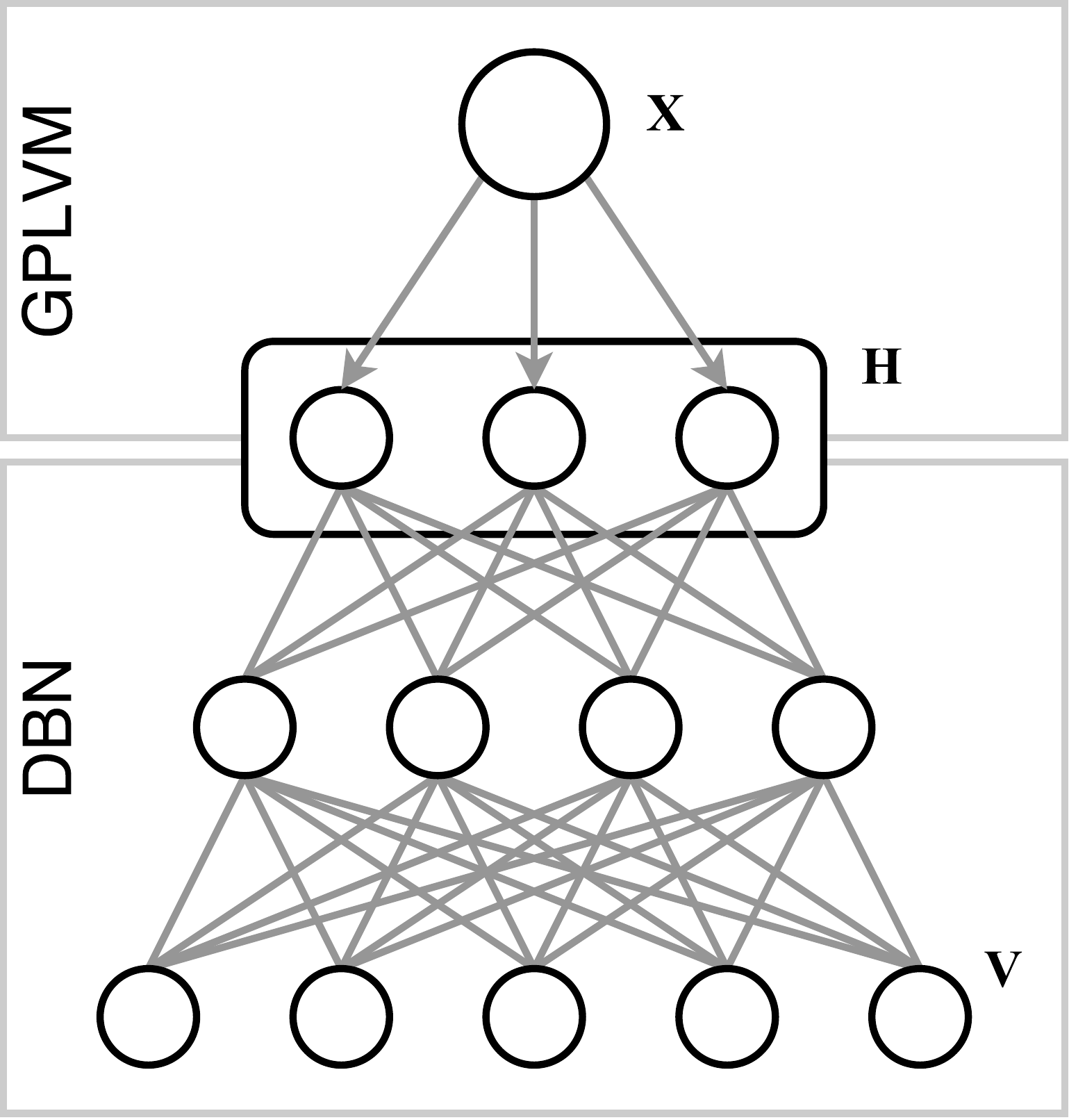}
        \label{fig:model-diagram}
    }\\[-5pt]%subfigure
   \caption{Graphical representations of the \subref{fig:rbm}~RBM, \subref{fig:dbn}~DBN and \subref{fig:sbm}~SBM. \subref{fig:model-diagram}~A graphical representation of our proposed GPDBN model where $\bm{X}$ represents the latent variables, $\bm{H}$ the Gaussian activations~\eqref{gaussian-activations}, and $\bm{V}$ the observed (data) space. \emph{(The SBM figure is taken from~\cite{eslami2013shape}.)}}
\label{fig:rbm-dbn-sbm}
\end{figure}

\boldparagraph{RBM}
The restricted Boltzmann machine (RBM), or Harmonium,~\cite{smolensky1986parallel} is a generative stochastic neural network that learns a probability distribution over a vector of random variables. The RBM is when stacked the basic the basic component of a deep belief network. The graphical model of the RBM is an undirected bipartite graph, consisting of a set of visible random variables (or units): $\bm{v}$, and a set of hidden units $\bm{h}$ (Fig.~\ref{fig:rbm}). Typically, all variables are binary (Bernoulli), taking on values from $\{0,1\}$.

The RBM model specifies a probability distribution over both the visible and hidden variables jointly as
\begin{equation}\label{gibbs_distr}
P(\bm{v}, \bm{h}) = \frac{1}{Z}\exp{(-E(\bm{v}, \bm{h}))}
\end{equation}
which defines a Gibbs distribution with energy function
\begin{equation}\label{rbm-energy}
E(\bm{v}, \bm{h}) = -\bm{v}^\top\bm{W}\bm{h} - \bm{b}^\top\bm{v} - \bm{c}^\top\bm{h}\ ,
\end{equation}
where $\bm{W}$, $\bm{b}$, $\bm{c}$ are the parameters of the model: $\bm{W}$ as a linear weight matrix and $(\bm{b},\bm{c})$ are bias vectors for the visible and hidden units respectively. The normalising constant $Z$ is the, computationally intractable, sum over all possible random vectors $\bm{v}$ and $\bm{h}$.

The bipartite structure of the model (\ie, the graph has no visible-visible or hidden-hidden connections, as shown in Fig.~\ref{fig:rbm}), affords efficient Gibbs sampling from the visible units given the hidden variables (or vice versa).
The conditional distribution of the hidden units given the visible ones, and vice versa, factorize as each set of variables are conditionally independent given the other:
\begin{equation}
P(\bm{h} \,|\, \bm{v}) = \textstyle\prod_{j=1}^H P(h_j \,|\, \bm{v}), \;
P(\bm{v} \,|\, \bm{h}) = \textstyle\prod_{i=1}^V P(v_i \,|\, \bm{h}) \label{rbm-conditional}\ .
\end{equation}
Replacing binary units with Gaussian units can be performed by modifying the energy function~\cite{hinton2012practical}.
Unfortunately, parameter learning is difficult since direct calculation of the gradients of the log likelihood w.r.t. the parameters requires the intractable computation of the normalising constant $Z$. In \emph{current} practice, the approximate maximum-likelihood contrastive divergence algorithm is used~\cite{carreira2005contrastive}.

\boldparagraph{DBN}
When multiple layers of RBMs are stacked on top of each other they form a deep belief network (Fig.~\ref{fig:dbn}). Hinton~\etal~\cite{hinton2006reducing} demonstrated that a DBN can be trained in a greedy fashion, layer by layer. Essentially, the samples (activations) from the hidden units of a trained layer are used as the data to train the next layer in the stack.
% CAN REMOVE THIS IF WE GET SQUISHED FOR SPACE
\boldparagraph{Sampling}
Sampling from an RBM proceeds by conditioning on some input data and performing a Gibbs sample for the hidden units. Subsequently, a Gibbs sample can be drawn for the visible units by conditioning the hidden units on this sample. This process is then repeated for a number of cycles.
Since a DBN is a stack of RBMs, this process has to be repeated for all layers; the output of one layer becomes the input to condition on for the next layer. In this way, an input data point can be propagated up and down the network.

\boldparagraph{Limitations}
Although a DBN is good at learning low-dimensional stochastic representations of high-dimensional data, it has three key drawbacks that we will address by combining the strengths of the DBN with a flexible non-parametric model in \S~\ref{model}:
\begin{enumerate}
\item It lacks a directed generative sampling process from a well defined latent representation. In order to generate a sample one must condition on some input data and propagate it through the network back and forth until a sample from the lowest layer is obtained.\\[-11pt]
\item There is no explicit representation of the uncertainty, instead this only arises implicitly through the propagation of point estimates (samples) at each layer.\\[-11pt]
\item A side effect of the conditional independence assumption of~\eqref{rbm-conditional} is that the correlations between the hidden units of the top layer of a DBN are not captured because each latent dimension is independent. Most importantly, a DBN does not, therefore, give any guarantee about learning a smooth latent space.
\end{enumerate}

%===========================================================
\subsection{GPLVM}
\label{gp}

\newcommand{\GP}{\mathrm{GP}}

The Gaussian Process Latent Variable Model (GPLVM)~\cite{lawrence2005probabilistic} learns a generative representation by placing a Gaussian process (GP) prior over the mapping from the latent to the observed data. This approach has the benefit that it is very easy to ensure a smooth mapping from the latent representations to the observed data. Further, due to the principled uncertainty propagation of the GP, all predictions will have an associated uncertainty.

In specific, each observed datapoint $\bm{y}_{n}$, $n \in [1,N]$, is assumed to be generated by a latent location $\bm{x}_{n}$ through a mapping $f$. Due to the marginalising property of a Gaussian, the predictive posterior over function values $\bm{f}^*$ at a test location $\bm{x}^*$ can be reached in closed form as,
\begin{gather}\label{noise-free-conditional}
p(\bm{f}^* \mid \bm{Y}, \bm{x}^*, \bm{X}) = \mathcal{N}(\bm{m}_{\GP}, \sigma^2_{\GP})\\
\bm{m}_{\GP} = k(\bm{x}^*, \bm{X}) [k(\bm{X}, \bm{X})]^{-1} \bm{Y}\label{gp-pred-mean}\\ %\bm{K}^{-1}
\sigma^2_{\GP} = k(\bm{x}^*, \bm{x}^*) - k(\bm{x}^*, \bm{X}) [k(\bm{X}, \bm{X})]^{-1} k(\bm{X}, \bm{x}^*)\label{gp-pred-var}\ ,
\end{gather}
where $k(\cdot,\cdot)$ is the covariance function specifying the Gaussian process and $\bm{X} = [\bm{x}_1,\dots,\bm{x}_N]^\top$.
We used the common  \emph{squared exponential} kernel
\begin{equation}\label{squared-exponential}
k(\bm{x}, \bm{x}') = \alpha^2\exp\left(-\frac{1}{2\ell^2}\left\lVert \bm{x} - \bm{x}' \right\rVert^2 \right) \ ,
\end{equation}
with hyperparameters $\alpha^{2}$ (signal variance) and $\ell$ (lengthscale), to ensure a smooth manifold.
Importantly, even though the function $f$ can be non-linear, the relationship between the predicted mean \eqref{gp-pred-mean} and the training data $\bm{Y}$ is linear. Due to this linearity, a GPLVM is inherently not suitable for modeling image data.

%===========================================================
\subsection{Shape Boltzmann Machine}
\label{sbm}

The Shape Boltzmann Machine (SBM)~\cite{eslami2013shape} is a specific architecture of the Boltzmann machine.
It consists of three layers: a rectangular layer of $N \times M$ visible units $\bm{v}$, and two layers of latent variables: $\bm{h}^{1}$ and $\bm{h}^{2}$.
Each hidden unit in $\bm{h}^1$ is connected only to one of the four subsets of visible units of $\bm{v}$ (Fig. \ref{fig:sbm}).
Each subset forms a rectangular patch and the weights of each patch (except the biases) are shared so that a patch effectively behaves as a local receptive field.
To avoid boundary inconsistencies, the patches are slightly overlapped (in Fig. \ref{fig:sbm}, the overlap has size $b$). Layer $\bm{h}^{2}$ is fully connected to $\bm{h}^{1}$.

% THIS CAN BE REMOVED IF NECESSARY
While the SBM offers improved generalization over a DBN with the same number of parameters, the SBM has a fixed structure which is not easily extended to more layers or patches. In contrast, a DBN, as a stack of simple RBMs, has a more generic and flexible structure which can be adapted easily and combined with other models. Furthermore, like the DBN, the SBM lacks of a proper generative process.

%===========================================================
\section{The GPDBN Model}
\label{model}

In our model, we connect a DBN and GPLVM so that the data space of the GPLVM corresponds the latent space of the DBN (Fig.~\ref{fig:model-diagram}) to obtain a model that can be optimized by minimizing a single objective function.

\boldparagraph{New Concrete Layers}
The uppermost hidden layer of the DBN has Gaussian units to interface with the Gaussian likelihood of the GPLVM.
In the lower layers, we replace the standard binary units with a \emph{Concrete distribution}~\cite{maddison2017concrete}.
This is a continuous relaxation to discrete random variables, in our case, to the Bernoulli distribution.
This allows us to draw low bias samples, in an analogous manner to the reparameterization trick~\cite{Kingma2013}, using a function that is differentiable with respect to the model parameters,
\begin{equation}
\mathrm{Concrete}\left(p, u\right) = \mathrm{Sigmoid}\!\left(\textstyle\frac{1}{\lambda}\big(\log p -\log(1-p) + \log u - \log(1 - u) \big) \right)\ ,\label{concrete}
\end{equation}
where $p$ is the parameter of a Bernoulli distribution, $\lambda$ is a scaling factor, which we fix to $0.1$, and $u$ is a uniform sample from $[0,1]$.

\boldparagraph{Learning}
Given a dataset $\mathcal{D} = \{\bm{t}_{n}\}_{n=1}^N$, we train the model end-to-end by minimizing the following objective function jointly with respect to all the parameters and the matrix of latent points $\bm{X}$ (omitted from the notation to avoid clutter):%
\begin{equation}\label{objective}
{\scriptsize{%
L = \textstyle\sum_{n=1}^N \underbrace{\left(\bm{t}_{n}\log(\bm{s}_{n}) + (1 - \bm{t}_{n})\log(1 - \bm{s}_{n}) \right)}_{\text{data term}}
+ \frac{1}{2} \underbrace{\mathrm{Tr}\!\left[\bm{K}^{-1}\bm{H} \bm{H}^\top\right]}_{\text{joint term}}
+ \frac{D}{2} \underbrace{\log |\bm{K}|}_{\text{\shortstack{complexity\\term}}} + \underbrace{||\bm{X}||^2}_{\text{\shortstack{prior\\term}}}\ .%
}}
\end{equation}
Here, $\bm{t}_{n}$ is a training datapoint, $\bm{s}_{n}$ is a sample from the model, $\bm{K} = k\big(\bm{X}, \bm{X}\big) + \sigma^{2}\bm{I}_N$ is the covariance matrix of the latent points and $D$ is the number of Gaussian units in the uppermost DBN layer (equal to the dimension of the GPLVM output space). We use a standard Gaussian as the prior on $\bm{X}$.
The variance of the noise parameter is $\sigma^2$ and $\bm{I}_N$ is an $N \times N$ identity matrix.

\newcommand{\DBN}{\mathrm{DBN}}

To join the two models, the $N \times D$ matrix of activations $\bm{H}$, from the Gaussian units, is defined as:
\begin{equation}\label{gaussian-activations}
\bm{H} = \bm{A} + \bm{\sigma}^{\GP} \otimes \bm{\sigma}^{\DBN} \odot \bm{\mathcal{E}}\ ,
\end{equation}
where $\bm{A} = [\bm{m}^{\GP}_1, \dots, \bm{m}^{\GP}_N]^{\top}$ is a matrix in which each row is the mean output of the Gaussian units corresponding to each input training datapoint.
This is combined with $\bm{\sigma}^{\GP}$, the $N \times 1$ vector of predictive standard deviations from the GPLVM
, and $\bm{\sigma}^{\DBN}$, the $1 \times D$ vector of standard deviation parameters of the Gaussian units.
Note that $\otimes$ is an outer product, and $\odot$ is an element-wise product.

The $\bm{H}$ matrix represents the observed data for the GPLVM and is updated at each training iteration by sampling $\bm{\mathcal{E}}$ a different $N \times D$ matrix of independent Gaussian noise, $\mathcal{E}_{n,d}\sim\mathcal{N}(0,1)$. This is a second application of the reparameterization trick.
 At each iteration, $\bm{H}$ is always normalized, to match our zero mean GP assumption, by subtracting its column-wise mean and dividing by $\bm{\sigma}^{\DBN}$.

\boldparagraph{Minibatches}
The objective \eqref{objective} can be evaluated on an uniformly drawn subset of data $\{\bm{t}_{b}\}_{b=1}^B$ yielding an estimator for the full objective,
\begin{align}\label{estimated_objective}\small
 L_{\text{batched}} &\simeq \textstyle\frac{N}{B} \textstyle\sum_{b=1}^B \big(\bm{t}_{b}\log(\bm{s}_{b}) + (1 - \bm{t}_{b})\log(1 - \bm{s}_{b})\big)
+ \frac{N}{2B} \mathrm{Tr}\left[\bm{K}_{B}^{-1}\bm{H}_{B} \bm{H}_{B}^\top\right]\nonumber\\[3pt]
&\qquad + \textstyle\frac{ND}{2B}\log |\bm{K}_{B}| + \frac{N}{B}||\bm{X}_{B}||^2 \ ,
\end{align}
where $\bm{H}_{B}$ and $\bm{K}_{B}$ corresponds to $\bm{H}$ and $\bm{K}$ evaluated on the subset $\bm{X}_{B}$ of $\bm{X}$.
Using this estimator the model can be optimised using mini-batching to scale linearly to larger datasets.
We note that the matrix inversion does introduce bias into the estimator; empirical results suggest this is small and removing it is a topic for future work.

\boldparagraph{Scaling via Convolutional Architecture} When defining the likelihood directly over the pixels, the fully-connected conditional independence of the RBM layers limits scalability in terms of image size.
This can be circumvented by adding convolution and deconvolution steps to replace the dense matrix product in~\eqref{rbm-energy} in the lower layers.

\boldparagraph{Sampling}
A sample $\bm{s}_{n}$ from the model is drawn by first generating a hidden sample $\bm{h}_{n}$ %(of size $1 \times D$,
from latent point $\bm{x}_{n}$:
\begin{equation}\label{h_n}
\bm{h}_{n}(\bm{x}) = (\bm{m}^{\GP}_{n} + \sigma^{\GP}_{n} \times \bm{\epsilon}_{n}) \odot \bm{\sigma}^{\DBN} + \bm{h}_{\mu}\ ,
\end{equation}
using $\bm{m}^{\GP}_{n}$ and $\sigma^{\GP}_{n}$ as the predictive mean and standard deviation of the GPLVM given latent point $\bm{x}_{n}$. This is combined with a sample $\bm{\epsilon}_{n}$, a $1 \times D$ vector of spherical Gaussian noise. The term $\bm{h}_{\mu}$ is the mean vector that is subtracted from $\bm{H}$ in the normalization step.
The sample $\bm{h}_{n}$ is then propagated down through the DBN, sampling layer-by-layer, to give an output sample $\bm{s}_{n}$.

\boldparagraph{Prediction and Projection}
Since we have a simple sampling process, we can propagate uncertainty for our predictions by taking the empirical mean of a set of $J$ samples from the model as $\bm{s}_* = \frac{1}{J} \sum_{j}^{J} \bm{s}_j \!\big( \,\bm{h}_j(\bm{x}_*) \big)$ for the latent location $\bm{x}^{*}$.
Since we can efficiently take gradients through the sampling process, we can project new observations into the latent space by minimizing the reprojection error w.r.t. the latent locations for predictions from a set of random starting locations in the manifold.

\boldparagraph{Interpretation}
We note that the objective~\eqref{objective} consists of terms in contrast with each other. The first encodes a \emph{data} term that ensures the observed data is well represented by the model. The third provides a \emph{complexity} term that encourages a simple (low complexity) latent space $\bm{X}$ through the covariance matrix $\bm{K}$ to prevent overfitting.

The second term ``glues'' the two models together by ensuring that the covariance matrix $\bm{K}$ is a good model of the covariance of the Gaussian units at the top of the DBN. This in turn, ensures that the DBN learns an appropriate network to give sensible Gaussian activations rather than the unconstrained binary activations from a normal DBN. The last term encodes a \emph{prior} which encourages the latent points to stay close to the origin.

The applications of the reparamerization trick ensures that efficient, low variance samples can be taken during training with gradients propagated throughout all parts of the network. The use of sampling and stochastic networks allows uncertainty to be propagated down through the entire model as well to ensure uncertainty is well quantified both at training and test time.

%===========================================================

\section{Experiments}
\label{experiments}

In keeping with previous work, we evaluated our models in terms of four experiments: (i)~\emph{Synthesis}, that is, generating samples that are plausible. (ii)~\emph{Representation and Generalisation}, demonstrating the ability to capture the variability of the silhouettes away from the training data. (iii)~\emph{Smoothness}, evaluating the quality of the learned latent space through interpolation; smooth trajectories in the latent space should produce smooth variations in the silhouette space. (iv)~\emph{Scaling}, evaluating how the model performs with respect to the size of the training dataset.

\boldparagraph{Our Models}
In the comparisons, our main model (which we will refer to as GPDBN) consists of a three-layer DBN plus a GPLVM layer connected as described in \S~\ref{model}. From the bottom (observed) to the top (hidden) layer the architecture consists of $200$ (Concrete units), $100$ (Concrete) and $50$ (Gaussian). The connected GPLVM layer has only $2$ latent dimensions for easy visualisation. The model is optimized jointly as described in~\S~\ref{model}.
Our second model, GPSBM, is similar to the GPDBN where the three-layer DBN has been replaced with an SBM architecture of~\cite{eslami2013shape} with hidden Concrete units in the bottom layer and hidden Gaussian units at the top.
We implemented all our models in the TensorFlow~\cite{Tensorflow} framework and trained using the Adam optimizer~\cite{AdamOpt}.

\boldparagraph{Baselines}
For comparison, we compared our models to size baselines: (i)~A vanilla GPLVM with $2$ latent dimensions. (ii)~GPLVMDT, a GPLVM operating on a signed distance function representation in a similar manner to~\cite{victor2012pwp3dsdf}; samples are obtained by thresholding through the hyperbolic tangent function. (iii)~The state-of-the-art ShapeOdds model~\cite{ShapeOdds}. (iv)~A DBN with binary units and the same architecture as our GPDBN. (v)~The SBM~\cite{eslami2013shape} model with binary units (trained layer by layer with contrastive divergence like the DBN) with the same architecture as our GPSBM. (vi)~The VAE~\cite{Kingma2013} model with the same architecture as our GPDBN (mirrored for the decoder) and $2$ latent dimensions. (vii)~An InfoGAN~\cite{chen2016infogan} with same %fully-connected network
architecture as the VAE and GPDBN (mirrored for the discriminator) and $2$ latent dimensions of structured noise.

\boldparagraph{Datasets}
In keeping with previous work, we trained the models on the Weizmann horse dataset~\cite{borenstein2004combining}, which consists of $328$ binary silhouettes of horses facing left. The limited number of training samples and the high variability in the position of heads, tails, and legs make this dataset difficult. We also trained the models on $300$  binary images from the Caltech101 dataset of motorbikes facing right~\cite{Caltech101}. All images in both datasets have been cropped and normalized to $32 \times 32$ pixels. The test datasets consisted of the challenging held-out data from~\cite{eslami2013shape}; an additional $14$ horses and $9$ motorbikes not contained in the training datasets.

\begin{figure}[t]
%\tiny
\centering
\setlength{\figwidth}{0.5\linewidth}
\subfigure[GPDBN horse manifold.]{%
\label{fig:gpdbn_horse_manifold}%
\begin{minipage}[c]{0.95\figwidth}
\centering
\includegraphics[width=\figwidth]{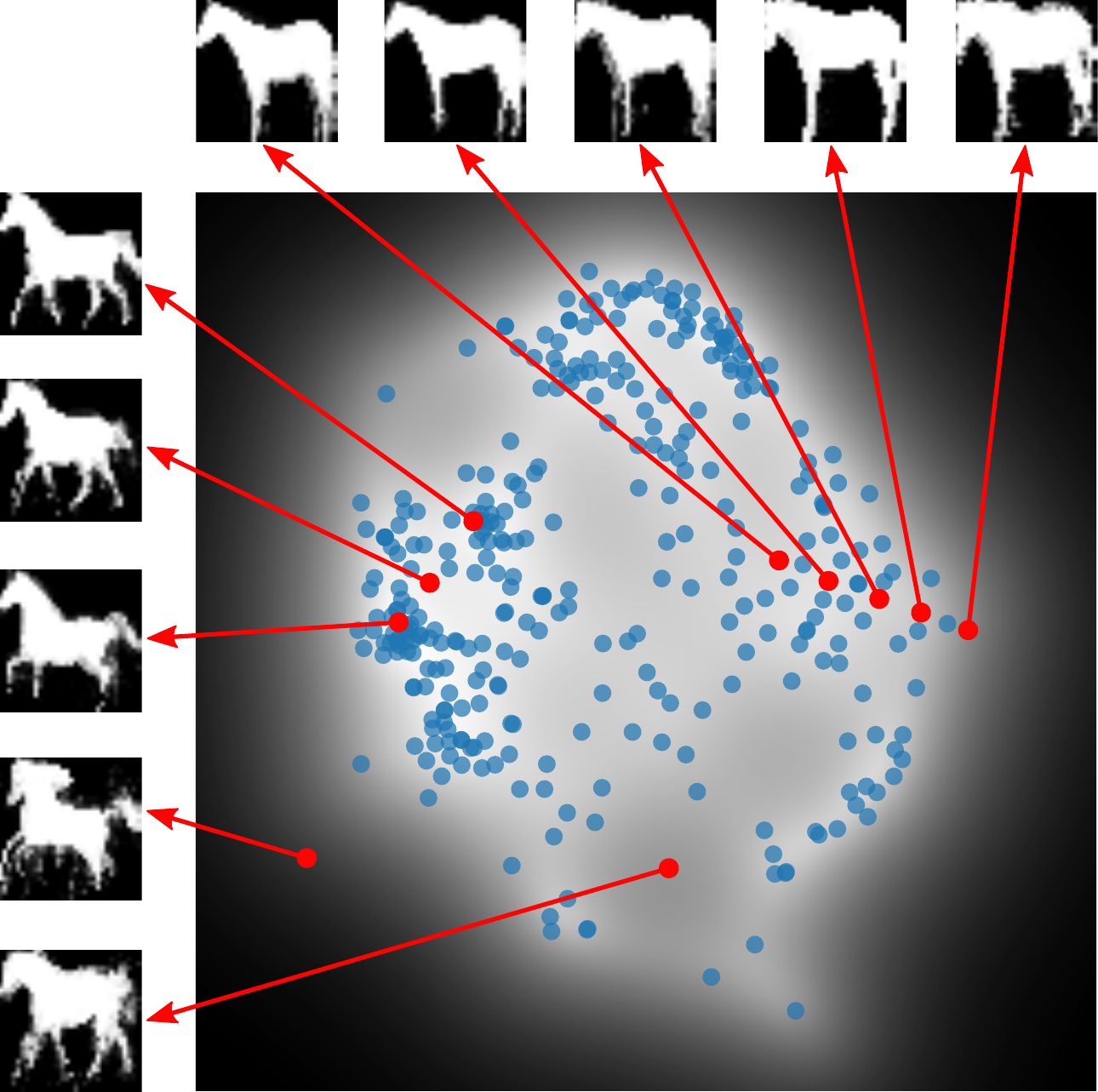}\vspace{2pt}
\end{minipage}
}\hfill%
\subfigure[Qualitative comparison of samples.]{%
\label{fig:realism}%
\begin{minipage}[c]{\figwidth}
\centering\scriptsize
\setlength{\tabcolsep}{3pt}
\begin{tabular}[t]{cccc}
{\scriptsize{GPLVM}} & \scriptsize{GPLVMDT} & \scriptsize{GPDBN} & \scriptsize{GPSBM} \\
\includegraphics[width=0.43in]{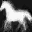} &
\includegraphics[width=0.43in]{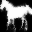} &
\includegraphics[width=0.43in]{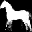} &
\includegraphics[width=0.43in]{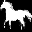}\\
\includegraphics[width=0.43in]{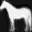} &
\includegraphics[width=0.43in]{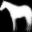} &
\includegraphics[width=0.43in]{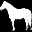} &
\includegraphics[width=0.43in]{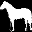}\\
\includegraphics[width=0.43in]{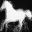} &
\includegraphics[width=0.43in]{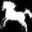} &
\includegraphics[width=0.43in]{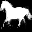} &
\includegraphics[width=0.43in]{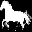}\\
\includegraphics[width=0.43in]{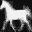} &
\includegraphics[width=0.43in]{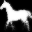} &
\includegraphics[width=0.43in]{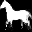} &
\includegraphics[width=0.43in]{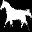}\\
\includegraphics[width=0.43in]{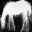} &
\includegraphics[width=0.43in]{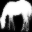} &
\includegraphics[width=0.43in]{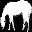} &
\includegraphics[width=0.43in]{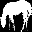}
\end{tabular}
\end{minipage}
}\\[-10pt]%
\caption{\subref{fig:gpdbn_horse_manifold}~Manifold learned by the GPDBN model on the Weizmann horse dataset. Moving over the manifold changes the pose of the horse with smooth paths in the manifold producing smooth transitions in silhouette pose. The heat map encodes the predictive variance of the model with darker regions indicating higher uncertainty and lower confidence in the silhouette estimates. \subref{fig:realism}~Qualitative comparison of silhouettes generated from low variance manifold areas by each of the models (images manually ordered by visual similarity).}
\label{fig:combined_manifold_realism}
\end{figure}

\boldparagraph{Synthesis}
Fig.~\ref{fig:gpdbn_horse_manifold}, shows the manifold learned by the GPDBN on the Weizmann horse dataset. Each blue point on the manifold represents the latent location corresponding to a training datapoint. The heat map is given by the log predictive variance \eqref{gp-pred-var} that encodes uncertainty in the latent space. The model is more likely to generate valid shapes from any location in the bright regions (\ie, low variance regions).

Unlike GP based models, a standard DBN (or the SBM) does not learn such a generative manifold. This implies, first of all, that a DBN does not allow us to sample ``from the top'' in a direct manner. Instead we must provide a test image to the visible units and condition on it before propagating it up and down the network for a few iterations to obtain a sample. Secondly, like the VAE and InfoGAN, a DBN does not provide information about how plausible a generated sample is.

A smooth generative manifold, such the one learned by our model in Fig.~\ref{fig:gpdbn_horse_manifold} is informative as it gives us an indication about where to sample from to get plausible silhouettes. Fig.~\ref{fig:realism} compares silhouettes generated by the models that allow sampling from the manifold.\footnote{When we show generated silhouettes from any model, we actually show grayscale images denoting pixel-wise probabilities of turning white rather than binary samples.}
We note that the GPLVM and GPLVMDT produce blurry images since the shapes present interpolation artifacts from the Gaussian likelihood. In contrast, the results from both the GPDBN and GPSBM are sharper.

\boldparagraph{Representation and Generalisation}
In the recent literature on shape modelling, quantitative results are reported in terms of the distance between the test data not seen by the model and the most likely prediction under the model.
For the models that can be sampled from, this amounts to finding the location on the manifold that most closely represents the test input (discussed for our model in \S~\ref{model}).
For the models that learn an explicit manifold we find the closest silhouette to a test silhouette $\bm{t}^{*}$ by minimising the following objective with respect to a latent location $\bm{x}^{*}$ on the manifold:
\begin{equation}\label{generalisation-objective}
L_{\text{proj}}(\bm{x}^{*}) = \textstyle\frac{1}{P}\textstyle\sum_{i=1}^V \left(\bm{t}^{*}\log(\bm{s}_{i}) + (1 - \bm{t}^{*})\log(1 - \bm{s}_{i}) \right) + \gamma\times \log(\sigma^2(\bm{x}^{*}))\ ,
\end{equation}
where we use $V$ samples to evaluate the cross entropy to the test silhouette.
The second term is the log predictive variance of the latent location $\bm{x}^{*}$ (as defined in Eq.\eqref{gp-pred-var}), this encourages the model to generate plausible silhouettes from the manifold. The scaling factor $\gamma$ ensures that the two term have approximatively the same scale.

\begin{figure}[t!]
\centering
\subfigure[Example results for projection onto manifold (20\% noise).]{%
\begin{minipage}[c]{0.72\linewidth}
    \tiny\centering
    \setlength{\figwidth}{0.27in}
    \begin{tabular}{ccccccccccc}
    %Unseen & 20\% Noise & DBN & SBM & VAE & InfoGAN & GPLVM & GPLVMDT & GPDBN & GPSBM \\
    unseen & 20\% & s.odds & dbn & sbm & vae & infogan & gpvlm & gplvmdt & gpdbn & gpsbm \\
    \includegraphics[width=\figwidth]{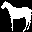} &
    \includegraphics[width=\figwidth]{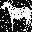} &
    \includegraphics[width=\figwidth]{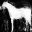} &
    \includegraphics[width=\figwidth]{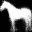} &
    \includegraphics[width=\figwidth]{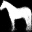} &
    \includegraphics[width=\figwidth]{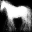} &
    \includegraphics[width=\figwidth]{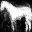} &
    \includegraphics[width=\figwidth]{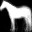} &
    \includegraphics[width=\figwidth]{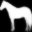} &
    \includegraphics[width=\figwidth]{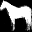} &
    \includegraphics[width=\figwidth]{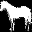}\\
    \includegraphics[width=\figwidth]{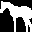} &
    \includegraphics[width=\figwidth]{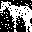} &
    \includegraphics[width=\figwidth]{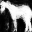} &
    \includegraphics[width=\figwidth]{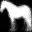} &
    \includegraphics[width=\figwidth]{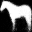} &
    \includegraphics[width=\figwidth]{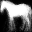} &
    \includegraphics[width=\figwidth]{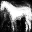} &
    \includegraphics[width=\figwidth]{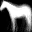} &
    \includegraphics[width=\figwidth]{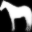} &
    \includegraphics[width=\figwidth]{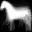} &
    \includegraphics[width=\figwidth]{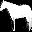}\\
    \includegraphics[width=\figwidth]{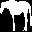} &
    \includegraphics[width=\figwidth]{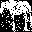} &
    \includegraphics[width=\figwidth]{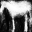} &
    \includegraphics[width=\figwidth]{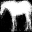} &
    \includegraphics[width=\figwidth]{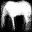} &
    \includegraphics[width=\figwidth]{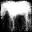} &
    \includegraphics[width=\figwidth]{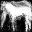} &
    \includegraphics[width=\figwidth]{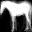} &
    \includegraphics[width=\figwidth]{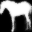} &
    \includegraphics[width=\figwidth]{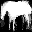} &
    \includegraphics[width=\figwidth]{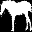}\\
    \includegraphics[width=\figwidth]{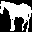} &
    \includegraphics[width=\figwidth]{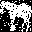} &
    \includegraphics[width=\figwidth]{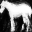} &
    \includegraphics[width=\figwidth]{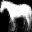} &
    \includegraphics[width=\figwidth]{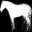} &
    \includegraphics[width=\figwidth]{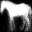} &
    \includegraphics[width=\figwidth]{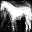} &
    \includegraphics[width=\figwidth]{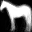} &
    \includegraphics[width=\figwidth]{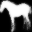} &
    \includegraphics[width=\figwidth]{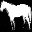} &
    \includegraphics[width=\figwidth]{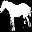}\\
    \includegraphics[width=\figwidth]{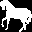} &
    \includegraphics[width=\figwidth]{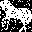} &
    \includegraphics[width=\figwidth]{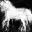} &
    \includegraphics[width=\figwidth]{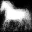} &
    \includegraphics[width=\figwidth]{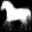} &
    \includegraphics[width=\figwidth]{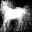} &
    \includegraphics[width=\figwidth]{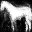} &
    \includegraphics[width=\figwidth]{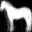} &
    \includegraphics[width=\figwidth]{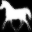} &
    \includegraphics[width=\figwidth]{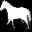} &
    \includegraphics[width=\figwidth]{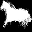}\\
    \includegraphics[width=\figwidth]{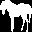} &
    \includegraphics[width=\figwidth]{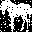} &
    \includegraphics[width=\figwidth]{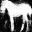} &
    \includegraphics[width=\figwidth]{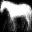} &
    \includegraphics[width=\figwidth]{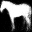} &
    \includegraphics[width=\figwidth]{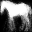} &
    \includegraphics[width=\figwidth]{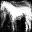} &
    \includegraphics[width=\figwidth]{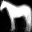} &
    \includegraphics[width=\figwidth]{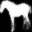} &
    \includegraphics[width=\figwidth]{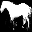} &
    \includegraphics[width=\figwidth]{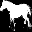}
    \end{tabular}
    \end{minipage}%
    \label{fig:generalisation_horses_20_noise}
    }\hfill%
    \subfigure[SSIM score (higher is better).]{%
\begin{minipage}[c]{0.26\linewidth}
\centering
\scalebox{0.8}{\small%
\begin{tabular}{lc}
Method & SSIM \\ \hline
ShapeOdds & $0.43 \pm 0.06$\\
DBN & $0.43 \pm 0.10$\\
SBM & $0.54 \pm 0.11$\\
VAE & $0.36 \pm 0.08$\\
InfoGAN & $0.27 \pm 0.06$\\
GPLVM & $0.48 \pm 0.07$\\
GPLVMDT & $0.54 \pm 0.09$\\[5pt]
\textbf{GPDBN} & $0.54 \pm 0.12$\\
\textbf{GPSBM} & $0.59 \pm 0.08$\\[10pt]
\end{tabular}}
\label{tab:results_horses_20_noise}
%\end{table*}
\end{minipage}
    }\\[-10pt]%
    \caption{Manifold projection from corrupted observations. \subref{fig:generalisation_horses_20_noise}~Test silhouettes (first column) are corrupted with 20\% salt and pepper noise (second column). The remaining columns show estimated silhouettes from each model. \subref{tab:results_horses_20_noise}~Mean and standard deviation of the SSIM score between silhouettes from each model against the original test data without noise.}
\label{fig:generalisation_horses_20_noise_and_table}
\end{figure}

\begin{figure*}[t!]
\centering
\subfigure[Example results for projection onto manifold (60\% noise).]{%
\begin{minipage}[c]{0.72\linewidth}
    \tiny\centering
    \setlength{\figwidth}{0.27in}
    \begin{tabular}{ccccccccccc}
    %Unseen & 60\% Noise & DBN & SBM & VAE & InfoGAN & GPLVM & GPLVMDT & GPDBN & GPSBM \\
    unseen & 60\% & s.odds & dbn & sbm & vae & infogan & gpvlm & gplvmdt & gpdbn & gpsbm \\
    \includegraphics[width=\figwidth]{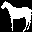} &
    \includegraphics[width=\figwidth]{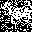} &
    \includegraphics[width=\figwidth]{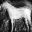} &
    \includegraphics[width=\figwidth]{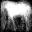} &
    \includegraphics[width=\figwidth]{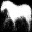} &
    \includegraphics[width=\figwidth]{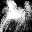} &
    \includegraphics[width=\figwidth]{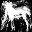} &
    \includegraphics[width=\figwidth]{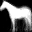} &
    \includegraphics[width=\figwidth]{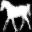} &
    \includegraphics[width=\figwidth]{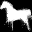} &
    \includegraphics[width=\figwidth]{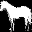}\\
    \includegraphics[width=\figwidth]{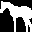} &
    \includegraphics[width=\figwidth]{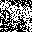} &
    \includegraphics[width=\figwidth]{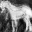} &
    \includegraphics[width=\figwidth]{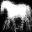} &
    \includegraphics[width=\figwidth]{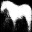} &
    \includegraphics[width=\figwidth]{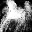} &
    \includegraphics[width=\figwidth]{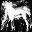} &
    \includegraphics[width=\figwidth]{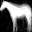} &
    \includegraphics[width=\figwidth]{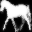} &
    \includegraphics[width=\figwidth]{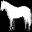} &
    \includegraphics[width=\figwidth]{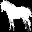}\\
    \includegraphics[width=\figwidth]{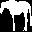} &
    \includegraphics[width=\figwidth]{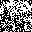} &
    \includegraphics[width=\figwidth]{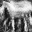} &
    \includegraphics[width=\figwidth]{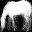} &
    \includegraphics[width=\figwidth]{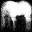} &
    \includegraphics[width=\figwidth]{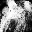} &
    \includegraphics[width=\figwidth]{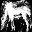} &
    \includegraphics[width=\figwidth]{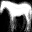} &
    \includegraphics[width=\figwidth]{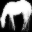} &
    \includegraphics[width=\figwidth]{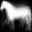} &
    \includegraphics[width=\figwidth]{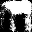}\\
    \includegraphics[width=\figwidth]{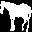} &
    \includegraphics[width=\figwidth]{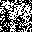} &
    \includegraphics[width=\figwidth]{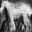} &
    \includegraphics[width=\figwidth]{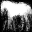} &
    \includegraphics[width=\figwidth]{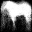} &
    \includegraphics[width=\figwidth]{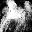} &
    \includegraphics[width=\figwidth]{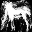} &
    \includegraphics[width=\figwidth]{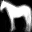} &
    \includegraphics[width=\figwidth]{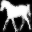} &
    \includegraphics[width=\figwidth]{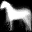} &
    \includegraphics[width=\figwidth]{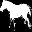}\\
    \includegraphics[width=\figwidth]{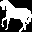} &
    \includegraphics[width=\figwidth]{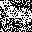} &
    \includegraphics[width=\figwidth]{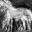} &
    \includegraphics[width=\figwidth]{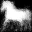} &
    \includegraphics[width=\figwidth]{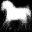} &
    \includegraphics[width=\figwidth]{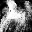} &
    \includegraphics[width=\figwidth]{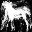} &
    \includegraphics[width=\figwidth]{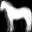} &
    \includegraphics[width=\figwidth]{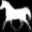} &
    \includegraphics[width=\figwidth]{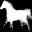} &
    \includegraphics[width=\figwidth]{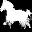}\\
    \includegraphics[width=\figwidth]{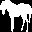} &
    \includegraphics[width=\figwidth]{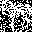} &
    \includegraphics[width=\figwidth]{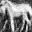} &
    \includegraphics[width=\figwidth]{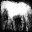} &
    \includegraphics[width=\figwidth]{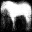} &
    \includegraphics[width=\figwidth]{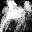} &
    \includegraphics[width=\figwidth]{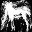} &
    \includegraphics[width=\figwidth]{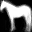} &
    \includegraphics[width=\figwidth]{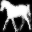} &
    \includegraphics[width=\figwidth]{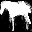} &
    \includegraphics[width=\figwidth]{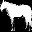}
    \end{tabular}
    \end{minipage}%
    \label{fig:generalisation_noisy_data}
    }\hfill%
    \subfigure[SSIM score (higher is better).]{%
\begin{minipage}[c]{0.26\linewidth}
\centering
\scalebox{0.8}{\small%
\begin{tabular}{lc}
Method & SSIM \\ \hline
ShapeOdds & $0.25 \pm 0.04$\\
DBN & $0.27 \pm 0.05$\\
SBM & $0.35 \pm 0.05$\\
VAE & $0.11 \pm 0.02$\\
InfoGAN & $0.18 \pm 0.03$\\
GPLVM & $0.44 \pm 0.07$\\
GPLVMDT & $0.37 \pm 0.03$\\[5pt]
\textbf{GPDBN} & $0.42 \pm 0.10$\\
\textbf{GPSBM} & $0.51 \pm 0.09$\\[10pt]
\end{tabular}}
\label{tab:results_noisy_data}
\end{minipage}
    }\\[-10pt]%
    \caption{Manifold projection from corrupted observations. \subref{fig:generalisation_noisy_data}~Test silhouettes (first column) are corrupted with 60\% salt and pepper noise (second column). The remaining columns show estimated silhouettes from each model. \subref{tab:results_noisy_data}~Mean and standard deviation of the SSIM score between silhouettes from each model against the original test data without noise. }
\label{fig:generalisation_noisy_data_and_table}
\end{figure*}

Samples for a DBN (or SBM) are usually generated by conditioning on an observed sample and propagating it through the network for several cycles, as described in \S~\ref{dbn}, with Gibbs samples taken after a burn in period. In our experiments, we fixed the conditioning on the test datapoint and averaged the results of a number of propagated samples through the model to prevent the sample chain from drifting away from the test data.

\boldparagraph{Projection under Noise}
To provide a challenging evaluation, we take unseen test data, corrupt it with noise and ask each models to find their most likely silhouette. Simply asking to reconstruct the test data would not be a sufficient evaluation since an identity mapping would be able to perform this task. Instead, we need the model to demonstrate that it can reject data that should not be in the trained model (the noise).
In Fig.~\ref{tab:results_horses_20_noise}, we report the results for our proposed model and the baseline methods. We use the Structured Similarity (SSIM)~\cite{ssim} metric (range [0,1] with high values better) with a small window size of $3$ to perform quantitative evaluations since it is known to outperform both cross-entropy and MSE as a perceptual metric.
A random sample of corresponding silhouettes for the horse dataset are provided in Fig.~\ref{fig:generalisation_horses_20_noise}.
We also test our model in a more challenging environment, Fig.~\ref{fig:generalisation_noisy_data_and_table}, where test data has been corrupted by significant noise.
The quantitative comparisons shown that our GPDBN and GPSBM models have captured a high quality probabilistic estimate of the data manifold while still preserving interpretability.

\begin{figure}[t]
\centering
\setlength{\tabcolsep}{0pt}
\begin{tabular}{C{0.12\linewidth}C{0.85\linewidth}}%{C{0.1\linewidth}C{0.9\linewidth}}
Dataset & \includegraphics[width=\linewidth]{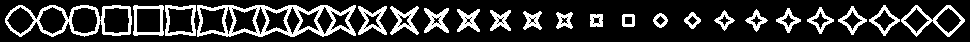}\\
\phantom{foo}\\[-5pt]
GPDBN&\includegraphics[width=\linewidth]{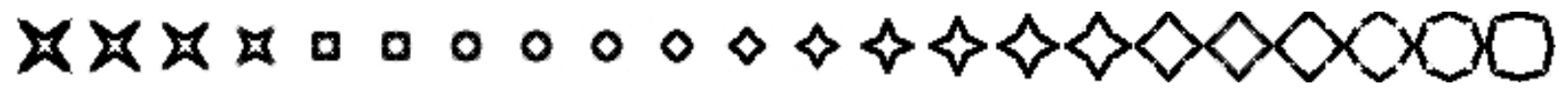}\\
VAE&\includegraphics[width=\linewidth]{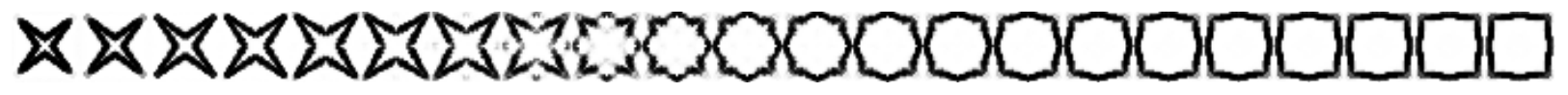}\\
InfoGAN&\includegraphics[width=\linewidth]{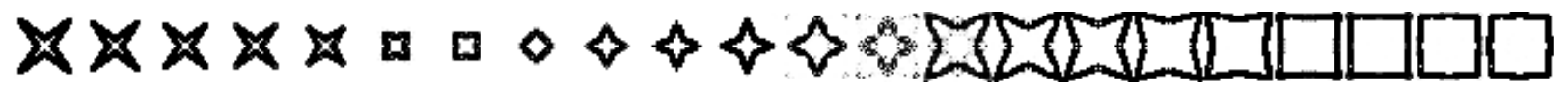}
\end{tabular}
\small
\setlength{\tabcolsep}{10pt}
\begin{tabular}{ccc}
%\hline
~GPDBN: $0.95 \pm 0.01$ ~&~ VAE: $0.87 \pm 0.03$ ~&~ InfoGAN: $0.93 \pm 0.06~$\\[-15pt]
%\hline
\end{tabular}
\caption{Example results of the interpolation test between two training points from the stars dataset. The top row shows the geodesic interpolation generated by the GPDBN. The middle and the last rows are the linear interpolation generated by the VAE and InfoGAN respectively.  The bottom row provides the mean and standard deviation of the SSIM score over $10$ interpolation experiments. (In this picture black and white are inverted respect to the training dataset.)}
\label{fig:geodesic_test}
\end{figure}

\boldparagraph{Interpolation Test}
We trained a GPDBN, VAE and InfoGAN models on a $30$ image dataset (which we call \emph{stars} dataset) generated from a \emph{known} 1-dimensional manifold using a simple script. The full dataset is displayed in the top row of Fig.~\ref{fig:geodesic_test}. The deterministically generated dataset allows us to  determine quantitatively whether interpolations in the latent space are representative of the true data distribution.
The middle rows of Fig.~\ref{fig:geodesic_test} show the model outputs for the interpolation between two latent points corresponding to a four-pointed \emph{star} (leftmost sample) and a \emph{square} (rightmost sample). The uncertainty information of the GPDBN allows us to go from one point to the other passing through low-variance regions by following a geodesic~\cite{Tosi:2014tt}. We can see that the GPDBN produces smoothly varying shapes of high quality that reflect the true manifold. In contrast, the VAE and InfoGAN results do not smoothly follow the true manifold and contain some erroneous interpolants that are not part of the true distribution; this is supported by the quantitative results that measure the quality of the samples to the true data using SSIM. The ability to exploit variance information in the GPDBN is clearly an advantage over the VAE and InfoGAN where the absence of direct access to the latent predictive posterior distribution prevents easy access to geodesics. Further demonstrations of the smoothness are available in supplementary material.

\begin{figure}[t]
    \centering
    \includegraphics[width=0.5\linewidth]{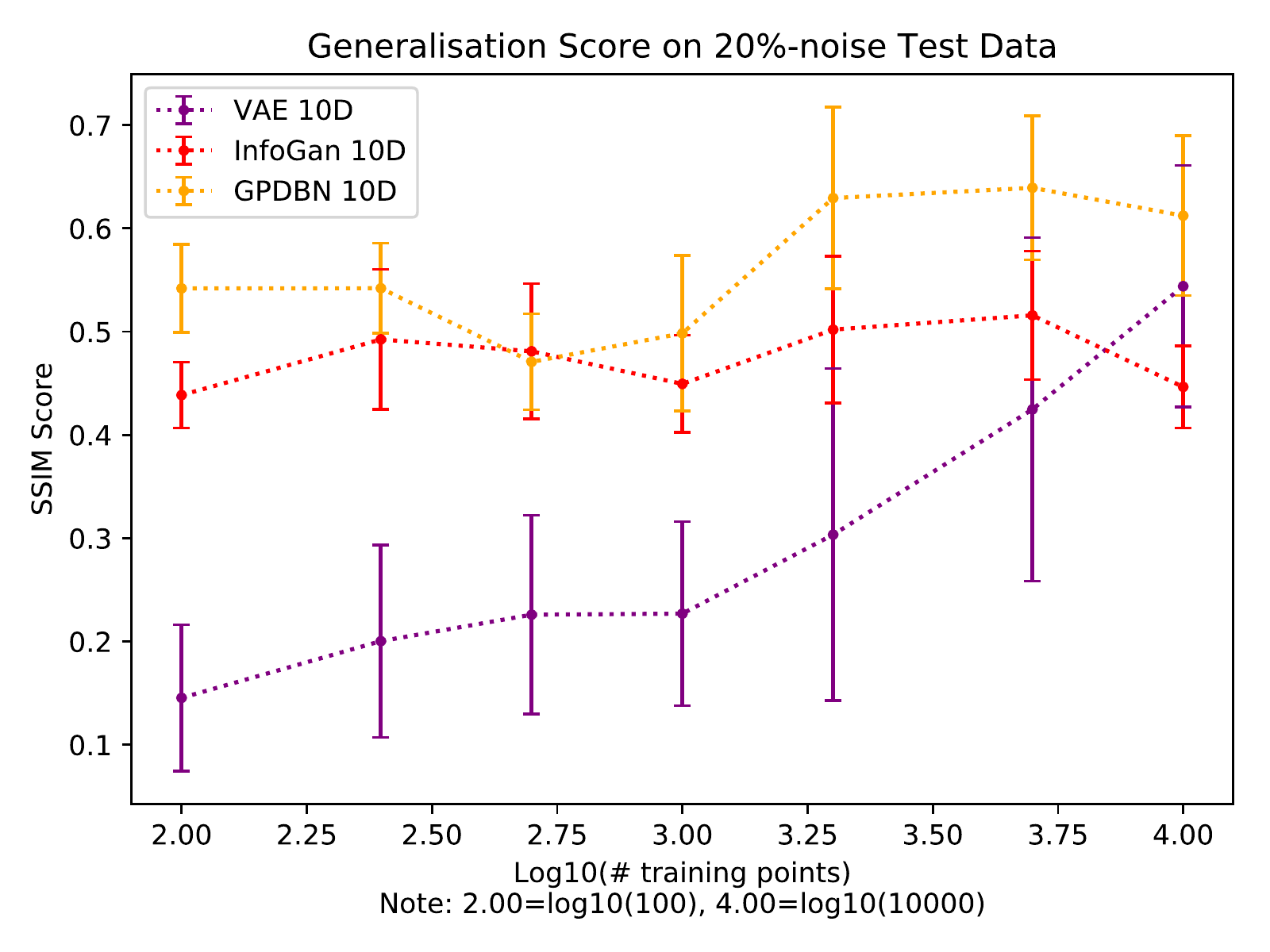}\\[-5pt]
    \caption{Graph showing the SSIM score of the output of the GPDBN, InfoGAN and VAE models against the test data without noise as the training dataset size increases from $100$ to $10000$ points. A higher score is better.}
    \label{fig:generalisation_graph}
\end{figure}

\begin{figure}[t]
    \centering
    \small
    \includegraphics[width=0.4\linewidth]{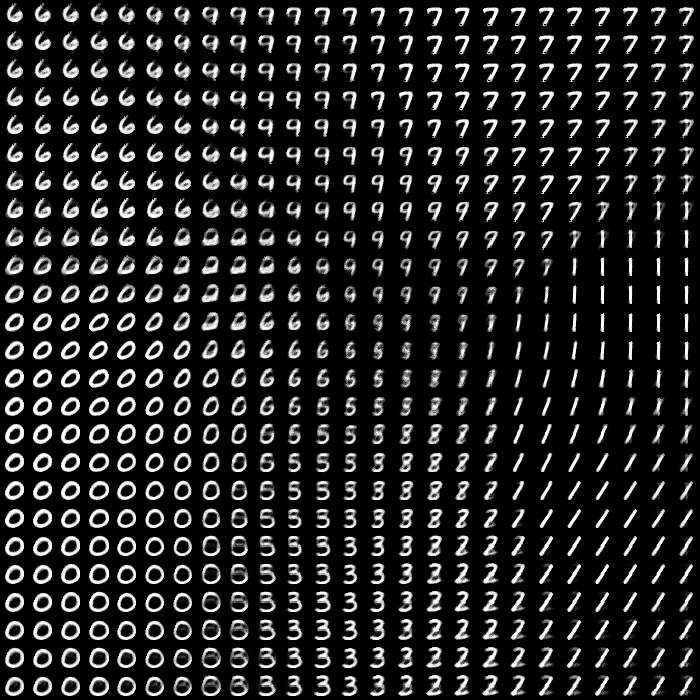}
    \includegraphics[width=0.4\linewidth]{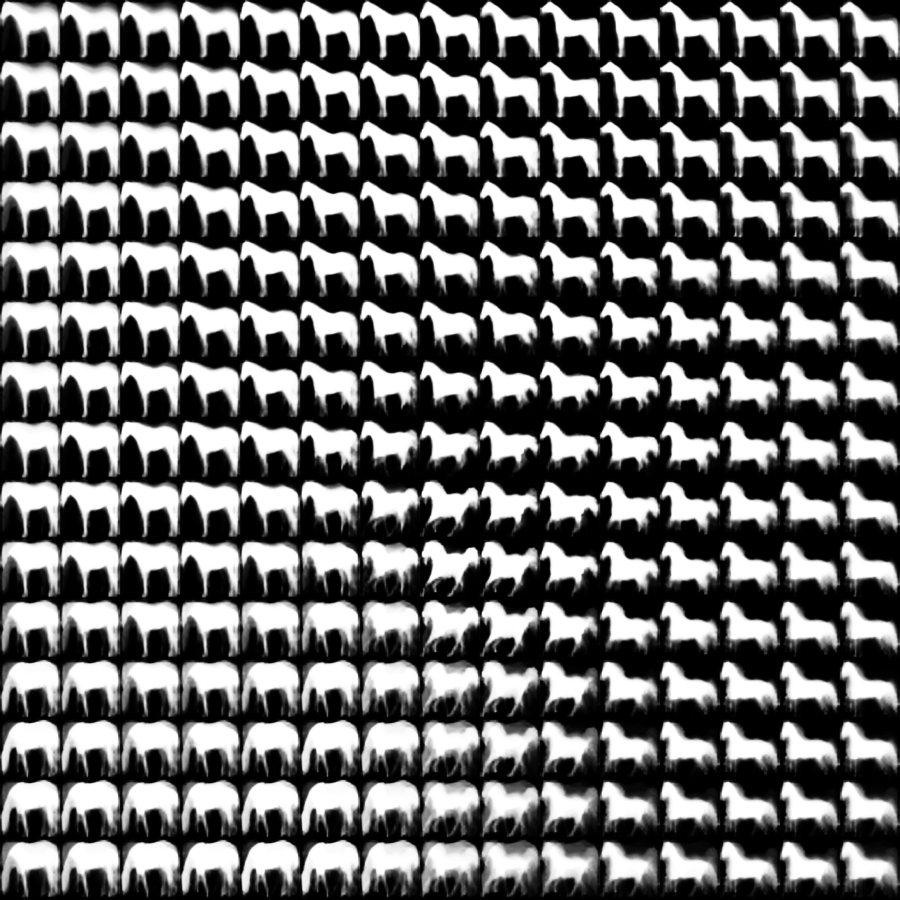}
    \caption{
    The model can be made to scale to a large number of datapoints by optimizing the objective using mini-batching\ \eqref{estimated_objective}.
    Furthermore, scalability in the size of images can be obtained by adding upscaling and convolutions in the lowermost layer.
    Left: GPDBN trained on MNIST comprised of 60,000 28x28 images.
    Right: GPDBN trained on Weizmann Horses comprised of 328 300x300 images.
    }
    \label{fig:scalability}
\end{figure}

\boldparagraph{Scaling Experiments}
In Fig.~\ref{fig:generalisation_graph} we compare the performance of the GPDBN, InfoGAN and VAE models as the size of the training dataset increases; here we use the standard MNIST digit dataset. We used a 10-dimensional latent space for all of the three models to account for the larger quantity of data. Similarly to the experiments in Figs. \ref{fig:generalisation_horses_20_noise_and_table} and \ref{fig:generalisation_noisy_data_and_table}, we took 30 random images from the MNIST test data, add 20\% salt-and-pepper noise, and calculated the SSIM score between the output of the models and the test data without noise. We plotted the score against dataset size (in log scale). We can see that the GPDBN model is able to capture a high quality model of the data manifold even from small datasets; for example, it achieves the same quality as a VAE trained on 10,000 images using only 100. We argue that the propagation of uncertainty throughout the model provides the advantage over both the VAE and InfoGAN which are both trained with only maximum likelihood approaches.

In Fig.~\ref{fig:scalability} we provide results that demonstrate that our approach also overcomes scaling issues normally present in GP models and DBNs. Firstly, we show training on the 60,000 MNIST images via our proposed mini-batching approach. In addition, we also show the manifold for higher resolution images from the horse dataset ($300\times300$). By using convolutional architectures, we can scale the number of parameters in an identical manner to convolutional feed-forward networks and our concrete layers allow us to train from random weight initialisation using back propagation without the need to use slow contrastive divergence. With both these approaches we still maintain our full uncertainty model so the same model can perform well with small and large datasets.

%===========================================================
\section{Conclusion}
\label{conclusion}
We have presented the GPDBN, a model that combines the properties of a smooth, interpretable low-dimensional latent representation with a data specific non-Gaussian likelihood function (for silhouette images). The model fully propagates and captures uncertainty in its estimates, it is trained end to end with the same complexity as a standard feed-forward neural network by minimising a single objective function, and is able to learn from very little data as well as scaling to larger datasets linearly by using mini-batching. We have shown both quantitatively and qualitatively that our model performs on par with the best shape models while at the same time introducing a smooth and low-dimensional latent representation with associated uncertainty that facilitates easy synthesis of data.

\boldparagraph{Acknowledgements}
This work was supported by the EPSRC CAMERA (EP/M023281/1) grant and the Royal Society.

%===========================================================

% \section{First Section}
% \subsection{A Subsection Sample}
% Please note that the first paragraph of a section or subsection is
% not indented. The first paragraph that follows a table, figure,
% equation etc. does not need an indent, either.
%
% Subsequent paragraphs, however, are indented.
%
% \subsubsection{Sample Heading (Third Level)} Only two levels of
% headings should be numbered. Lower level headings remain unnumbered;
% they are formatted as run-in headings.
%
% Headings should be capitalized
% (i.e., nouns, verbs, and all other words
% except articles, prepositions, and conjunctions should be set with an
% initial capital) and should,
% with the exception of the title, be aligned to the left.
% Words joined by a hyphen are subject to a special rule. If the first
% word can stand alone, the second word should be capitalized.
%
% \paragraph{Sample Heading (Fourth Level)}
% The contribution should contain no more than four levels of
% headings. Table~\ref{tab1} gives a summary of all heading levels.

\FloatBarrier

\bibliographystyle{splncs04}
\bibliography{references}

\begin{thebibliography}{10}
\providecommand{\url}[1]{\texttt{#1}}
\providecommand{\urlprefix}{URL }
\providecommand{\doi}[1]{https://doi.org/#1}

\bibitem{Tensorflow}
Abadi, M., et~al.: {{TensorFlow}: Large-Scale Machine Learning on Heterogeneous
  Systems} (2015), software available from tensorflow.org

\bibitem{bengio2007scaling}
Bengio, Y., LeCun, Y., et~al.: {Scaling learning algorithms towards AI}.
  Large-scale kernel machines  \textbf{34}(5),  1--41 (2007)

\bibitem{CampbellSIGGRAPH14}
Campbell, N.D.F., Kautz, J.: {Learning a manifold of fonts}. ACM Transactions
  on Graphics (SIGGRAPH)  \textbf{33}(4), ~91 (2014)

\bibitem{carreira2005contrastive}
Carreira{-}Perpi{\~{n}}{\'{a}}n, M.{\'{A}}., Hinton, G.E.: {On Contrastive
  Divergence Learning}. In: Cowell, R.G., Ghahramani, Z. (eds.) Proceedings of
  the Tenth International Workshop on Artificial Intelligence and Statistics,
  {AISTATS} 2005, Bridgetown, Barbados, January 6-8, 2005. Society for
  Artificial Intelligence and Statistics (2005)

\bibitem{chen2016infogan}
Chen, X., Duan, Y., Houthooft, R., Schulman, J., Sutskever, I., Abbeel, P.:
  {InfoGAN: Interpretable Representation Learning by Information Maximizing
  Generative Adversarial Nets}. In: Lee, D.D., Sugiyama, M., von Luxburg, U.,
  Guyon, I., Garnett, R. (eds.) Advances in Neural Information Processing
  Systems 29: Annual Conference on Neural Information Processing Systems 2016,
  December 5-10, 2016, Barcelona, Spain. pp. 2172--2180 (2016)

\bibitem{damianou2013deep}
Damianou, A.C., Lawrence, N.D.: {Deep Gaussian Processes}. In: Proceedings of
  the Sixteenth International Conference on Artificial Intelligence and
  Statistics, {AISTATS} 2013, Scottsdale, AZ, USA, April 29 - May 1, 2013.
  {JMLR} Workshop and Conference Proceedings, vol.~31, pp. 207--215. JMLR.org
  (2013)

\bibitem{statModelsOfShape}
Davies, R., Twining, C., Taylor, C.: {Statistical models of shape: Optimisation
  and evaluation}. Springer Science \& Business Media (2008)

\bibitem{ShapeOdds}
Elhabian, S.Y., Whitaker, R.T.: {ShapeOdds: Variational Bayesian Learning of
  Generative Shape Models}. In: 2017 {IEEE} Conference on Computer Vision and
  Pattern Recognition, {CVPR} 2017, Honolulu, HI, USA, July 21-26, 2017. pp.
  2185--2196. {IEEE} Computer Society (2017)

\bibitem{eslami2012generative}
Eslami, S.M.A., Williams, C.K.I.: {A Generative Model for Parts-based Object
  Segmentation}. In: Bartlett, P.L., Pereira, F.C.N., Burges, C.J.C., Bottou,
  L., Weinberger, K.Q. (eds.) Advances in Neural Information Processing Systems
  25: 26th Annual Conference on Neural Information Processing Systems 2012.
  Proceedings of a meeting held December 3-6, 2012, Lake Tahoe, Nevada, United
  States. pp. 100--107 (2012)

\bibitem{eslami2013shape}
Eslami, S.A., Heess, N., Williams, C.K., Winn, J.: {The shape boltzmann
  machine: a strong model of object shape}. International Journal of Computer
  Vision  \textbf{107}(2),  155--176 (2014)

\bibitem{goodfellow2014gan}
Goodfellow, I.J., Pouget{-}Abadie, J., Mirza, M., Xu, B., Warde{-}Farley, D.,
  Ozair, S., Courville, A.C., Bengio, Y.: {Generative Adversarial Nets}. In:
  Ghahramani, Z., Welling, M., Cortes, C., Lawrence, N.D., Weinberger, K.Q.
  (eds.) Advances in Neural Information Processing Systems 27: Annual
  Conference on Neural Information Processing Systems 2014, December 8-13 2014,
  Montreal, Quebec, Canada. pp. 2672--2680 (2014)

\bibitem{hinton2012practical}
Hinton, G.E.: {A Practical Guide to Training Restricted Boltzmann Machines}.
  In: Montavon, G., Orr, G.B., M{\"{u}}ller, K. (eds.) Neural Networks: Tricks
  of the Trade - Second Edition, Lecture Notes in Computer Science, vol.~7700,
  pp. 599--619. Springer (2012)

\bibitem{hinton2006reducing}
Hinton, G.E., Salakhutdinov, R.R.: {Reducing the dimensionality of data with
  neural networks}. science  \textbf{313}(5786),  504--507 (2006)

\bibitem{AdamOpt}
Kingma, D.P., Ba, J.: {Adam: A Method for Stochastic Optimization}. In:
  International Conference on Learning Representations ({ICLR}) (2014)

\bibitem{Kingma2013}
Kingma, D.P., Welling, M.: {Auto-Encoding Variational Bayes}. In: International
  Conference on Learning Representations ({ICLR}) (2013)

\bibitem{deepPartBasedGenerativeShape}
Kirillov, A., Gavrikov, M., Lobacheva, E., Osokin, A., Vetrov, D.P.: {Deep
  Part-Based Generative Shape Model with Latent Variables}. In: Wilson, R.C.,
  Hancock, E.R., Smith, W.A.P. (eds.) Proceedings of the British Machine Vision
  Conference 2016, {BMVC} 2016, York, UK, September 19-22, 2016. {BMVA} Press
  (2016)

\bibitem{lawrence2005probabilistic}
Lawrence, N.D.: {Probabilistic Non-linear Principal Component Analysis with
  Gaussian Process Latent Variable Models}. Journal of Machine Learning
  Research  \textbf{6},  1783--1816 (2005)

\bibitem{lawrence2006local}
Lawrence, N.D., Candela, J.Q.: {Local distance preservation in the {GP-LVM}
  through back constraints}. In: Cohen, W.W., Moore, A. (eds.) Machine
  Learning, Proceedings of the Twenty-Third International Conference {(ICML}
  2006), Pittsburgh, Pennsylvania, USA, June 25-29, 2006. {ACM} International
  Conference Proceeding Series, vol.~148, pp. 513--520. {ACM} (2006)

\bibitem{Caltech101}
Li, F., Fergus, R., Perona, P.: {Learning Generative Visual Models from Few
  Training Examples: An Incremental Bayesian Approach Tested on 101 Object
  Categories}. In: {IEEE} Conference on Computer Vision and Pattern Recognition
  Workshops, {CVPR} Workshops 2004, Washington, DC, USA, June 27 - July 2,
  2004. p.~178. {IEEE} Computer Society (2004)

\bibitem{Rotopp2016}
Li, W., Viola, F., Starck, J., Brostow, G.J., Campbell, N.D.F.: {Roto++:
  Accelerating Professional Rotoscoping using Shape Manifolds}. ACM
  Transactions on Graphics (SIGGRAPH)  \textbf{35}(4), ~62 (2016)

\bibitem{maddison2017concrete}
Maddison, C.J., Mnih, A., Teh, Y.W.: {The Concrete Distribution: {A} Continuous
  Relaxation of Discrete Random Variables}. vol. abs/1611.00712 (2016)

\bibitem{victor2012pwp3dsdf}
Prisacariu, V.A., Reid, I.D.: {PWP3D: Real-time segmentation and tracking of 3D
  objects}. International journal of computer vision  \textbf{98}(3),  335--354
  (2012)

\bibitem{victor2011nonlinearshapemanifolds}
Prisacariu, V.A., Reid, I.D.: {Nonlinear shape manifolds as shape priors in
  level set segmentation and tracking}. In: The 24th {IEEE} Conference on
  Computer Vision and Pattern Recognition, {CVPR} 2011, Colorado Springs, CO,
  USA, 20-25 June 2011. pp. 2185--2192. {IEEE} Computer Society (2011)

\bibitem{borenstein2004combining}
Roy, A., Todorovic, S.: {Combining Bottom-Up, Top-Down, and Smoothness Cues for
  Weakly Supervised Image Segmentation}. In: 2017 {IEEE} Conference on Computer
  Vision and Pattern Recognition, {CVPR} 2017, Honolulu, HI, USA, July 21-26,
  2017. pp. 7282--7291. {IEEE} Computer Society (2017)

\bibitem{smolensky1986parallel}
Smolensky, P.: {Parallel Distributed Processing: Explorations in the
  Microstructure of Cognition, Vol. 1}. chap. Information Processing in
  Dynamical Systems: Foundations of Harmony Theory, pp. 194--281 (1986)

\bibitem{nonparametricGuidanceAutoEncoder}
Snoek, J., Adams, R.P., Larochelle, H.: {Nonparametric guidance of autoencoder
  representations using label information}. Journal of Machine Learning
  Research  \textbf{13},  2567--2588 (2012)

\bibitem{titsias2010bayesian}
Titsias, M.K., Lawrence, N.D.: Bayesian gaussian process latent variable model.
  In: Teh, Y.W., Titterington, D.M. (eds.) Proceedings of the Thirteenth
  International Conference on Artificial Intelligence and Statistics, {AISTATS}
  2010, Chia Laguna Resort, Sardinia, Italy, May 13-15, 2010. {JMLR}
  Proceedings, vol.~9, pp. 844--851. JMLR.org (2010)

\bibitem{Tosi:2014tt}
Tosi, A., Hauberg, S., Vellido, A., Lawrence, N.D.: {Metrics for Probabilistic
  Geometries}. In: Zhang, N.L., Tian, J. (eds.) Proceedings of the Thirtieth
  Conference on Uncertainty in Artificial Intelligence, {UAI} 2014, Quebec
  City, Quebec, Canada, July 23-27, 2014. pp. 800--808. {AUAI} Press (2014)

\bibitem{Vedaldi2015semanticpartsegmentation}
Tsogkas, S., Kokkinos, I., Papandreou, G., Vedaldi, A.: {Semantic Part
  Segmentation with Deep Learning}. arXiv preprint  \textbf{arXiv:1505.02438}
  (2015)

\bibitem{sketchPaper}
Turmukhambetov, D., Campbell, N.D.F., Goldman, D.B., Kautz, J.: {Interactive
  Sketch-Driven Image Synthesis}. Computer Graphics Forum  \textbf{34}(8),
  130--142 (2015)

\bibitem{ssim}
Wang, Z., Bovik, A.C., Sheikh, H.R., Simoncelli, E.P.: {Image quality
  assessment: from error visibility to structural similarity}. {IEEE} Trans.
  Image Processing  \textbf{13}(4),  600--612 (2004)

\bibitem{shapenets2015}
Wu, Z., Song, S., Khosla, A., Yu, F., Zhang, L., Tang, X., Xiao, J.: {3D
  ShapeNets: {A} deep representation for volumetric shapes}. In: {IEEE}
  Conference on Computer Vision and Pattern Recognition, {CVPR} 2015, Boston,
  MA, USA, June 7-12, 2015. pp. 1912--1920. {IEEE} Computer Society (2015)

\end{thebibliography}

%===========================================================
\hfill\break\newline\newline\newline\newline
{\noindent\bfseries\Large Supplementary Material}

%===========================================================
\section{Interactive Manifold Demonstration}

\begin{figure}[H]
\centering
\includegraphics[width=0.9\linewidth]{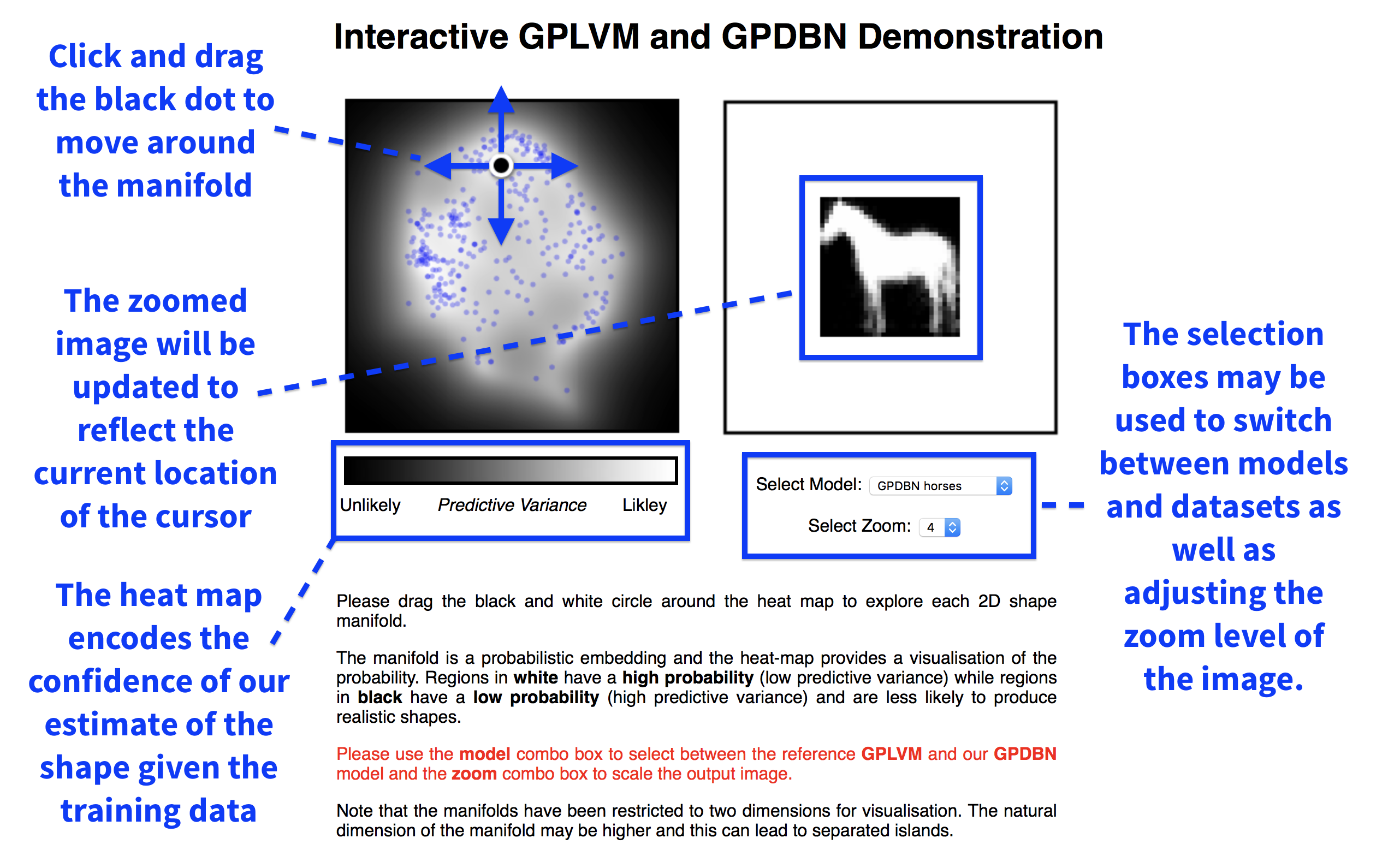}
\caption{The interactive manifold demonstration included as
%``\texttt{interactive\textunderscore demo.html}''
``\emph{interactive\_demo.html}''.
Please see the instructions for use shown in blue. While both the GPLVM and our GPDBN models provide interpretable, smooth manifolds, the Gaussian likelihood assumptions of the GPLVM lead to poor shape interpolations that are blurred and contain artifacts. Importantly, the heat map encodes the confidence in the estimated shapes via the predictive variance of the model.}
\label{fig:demo}
\end{figure}

\begin{figure}[H]
\begin{minipage}[t]{.47\textwidth}
\centering
\includegraphics[width=0.7\linewidth]{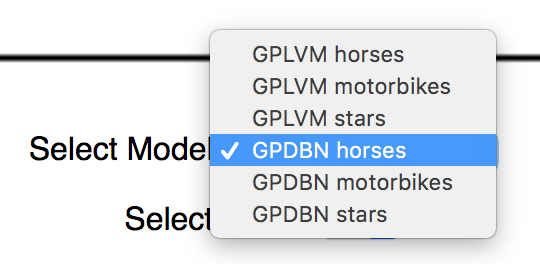}
\caption{Selecting the model combo box switches between models (GPLVM and GPDBN) and datasets (horses, motorbikes, cars, stars).}
\label{fig:demo_model_select}
\end{minipage}
\hfill
\noindent
\begin{minipage}[t]{.47\textwidth}
\centering
\includegraphics[width=0.7\linewidth]{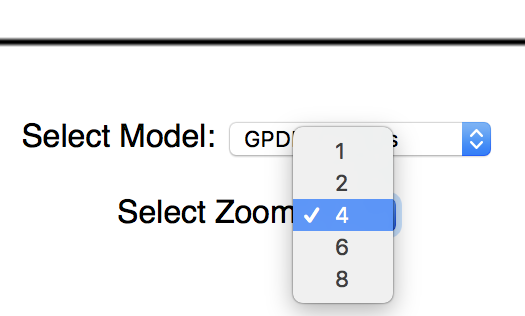}
\caption{Selecting the zoom combo box switches between different scales of the output image from $1$ (actual size) to $8$ times the actual size.}
\label{fig:demo_zoom_select}
\end{minipage}
\end{figure}

Please see the interactive manifold demonstration included with the supplemental material. The demonstration is a standalone application that can be viewed with a recent javascript, HTML5 compliant browser by loading the file
%``\texttt{interactive\textunderscore demo.html}''
``\emph{interactive\_demo.html}''. The data and rendering code is contained in the
%``\texttt{data}''
``\emph{data}''
directory is accessed through the standalone webpage. Figure~\ref{fig:demo} provides an illustration of what the page should look like with instructions for use in blue text. Figures~\ref{fig:demo_model_select} and \ref{fig:demo_zoom_select} show the options to select different models, datasets and zoom levels.

\section{Additional Results}

\begin{figure}[H]
\centering
\includegraphics[width=0.7\linewidth]{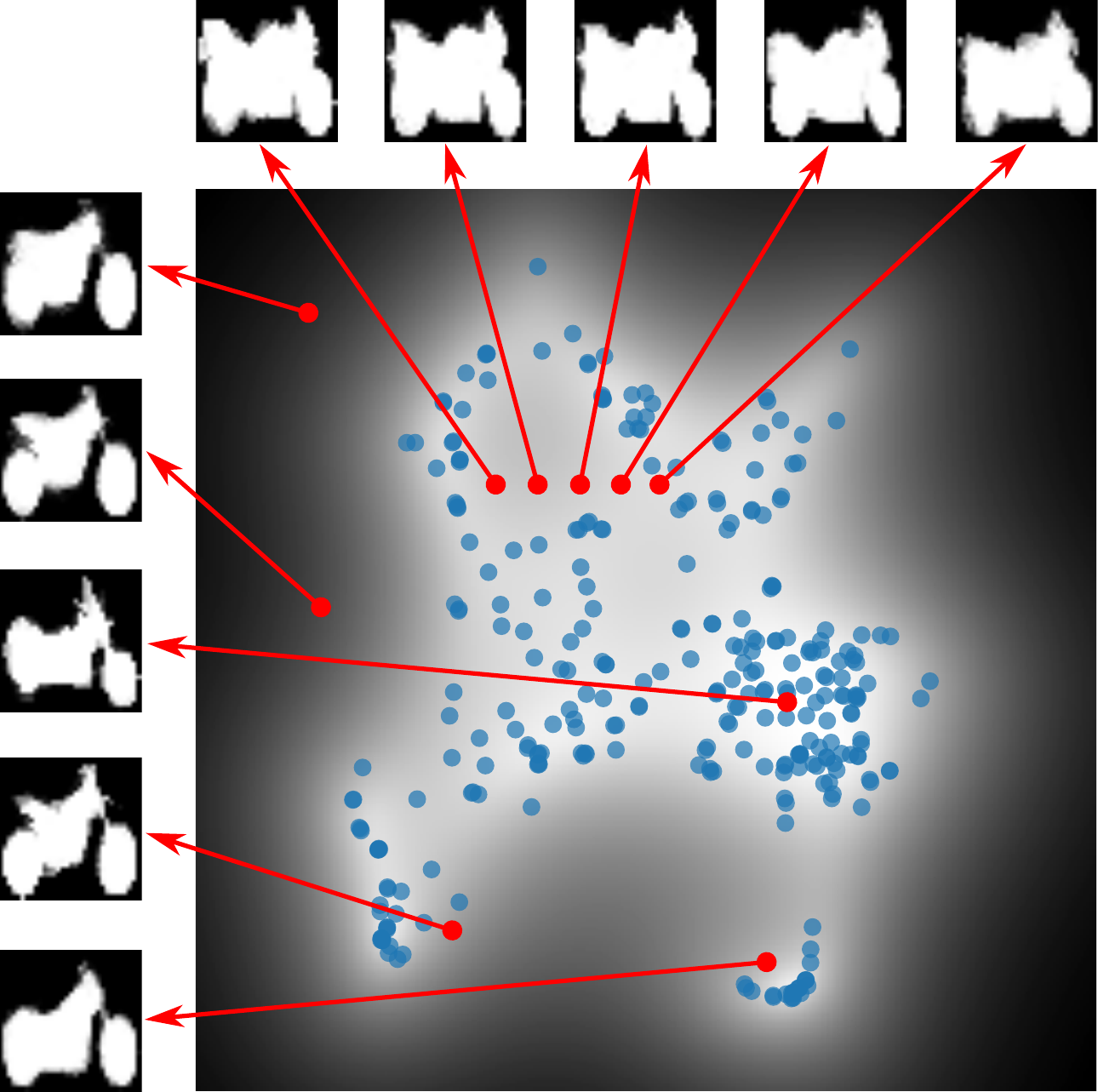}
\caption{Manifold learned by the GPDBN model on the motorbikes dataset. Moving over the manifold changes the shape of the motorbike producing smooth silhouette transitions. The heat map encodes the predictive variance of the model with darker regions indicating higher uncertainty and lower confidence in the silhouette estimates.}
\label{fig:gpdbn_motorbike_manifold}
\end{figure}

\begin{figure}[H]
\centering
\includegraphics[width=0.8\linewidth]{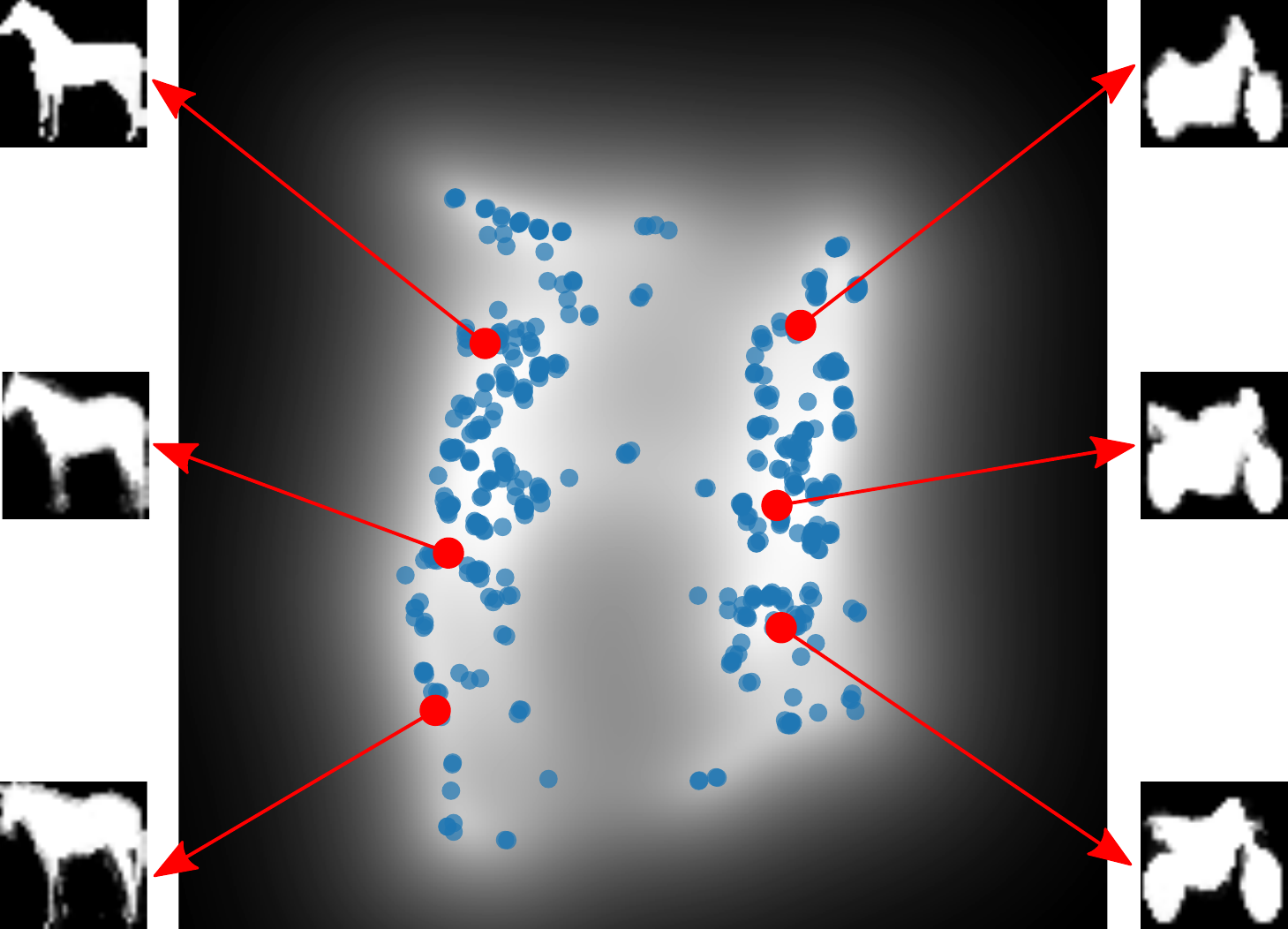}
\caption{Manifold learned by the GPDBN model trained on 250 horses + 250 bikes.}
\label{fig:gpdbn_horses_and_motorbikes_manifold}
\end{figure}

In Figure~\ref{fig:gpdbn_motorbike_manifold} we provide an illustration of the learned manifold for our GPDBN model on the  Caltech101 dataset of motorbikes facing right~\cite{Caltech101}.
We also trained a GPDBN on two types of data at the same time (horses and motorbikes). Fig.~\ref{fig:gpdbn_horses_and_motorbikes_manifold} shows what the manifold looks like (two separate clusters are clearly distinguishable).

\subsection{Additional Representation and Generalisation Results}

A fundamental ML problem is that there is no objective quantitative method of assessing unsupervised
learning models.
An identity function that simply outputs the test input would be able to achieve perfect reconstruction, however, under
such a model all silhouettes would be equally likely (even implausible ones); generalising to implausible shapes is not a property that we want.
In contrast, the predictive uncertainty of our GPDBN model tells us how plausible a generated silhouette is (a key feature).

Our proposal for a good quantitative assessment is to measure the reconstruction of the generated output
corresponding to a noisy version of the input.
An identity function would return the noisy version and be penalised for producing an implausible shape.
In contrast, a good model should return a projection to a plausible shape.
By using a noisy version the closest plausible shape should be the uncorrupted test image so we can compare to this to
quantify the performance.

\begin{figure}[H]
\centering
\subfigure[Example results for projection onto manifold]{%
\begin{minipage}[c]{0.72\linewidth}
    \tiny\centering
    \setlength{\figwidth}{0.27in}
    \begin{tabular}{cccccccccc}
    %Unseen & 20\% Noise & DBN & SBM & VAE & InfoGAN & GPLVM & GPLVMDT & GPDBN & GPSBM \\
    unseen & 10\% & dbn & sbm & vae & infogan & gpvlm & gplvmdt & gpdbn & gpsbm \\
    \includegraphics[width=\figwidth]{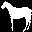} &
    \includegraphics[width=\figwidth]{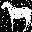} &
    \includegraphics[width=\figwidth]{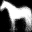} &
    \includegraphics[width=\figwidth]{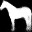} &
    \includegraphics[width=\figwidth]{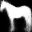} &
    \includegraphics[width=\figwidth]{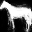} &
    \includegraphics[width=\figwidth]{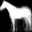} &
    \includegraphics[width=\figwidth]{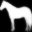} &
    \includegraphics[width=\figwidth]{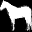} &
    \includegraphics[width=\figwidth]{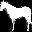}\\
    \includegraphics[width=\figwidth]{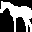} &
    \includegraphics[width=\figwidth]{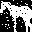} &
    \includegraphics[width=\figwidth]{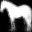} &
    \includegraphics[width=\figwidth]{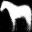} &
    \includegraphics[width=\figwidth]{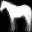} &
    \includegraphics[width=\figwidth]{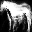} &
    \includegraphics[width=\figwidth]{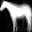} &
    \includegraphics[width=\figwidth]{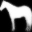} &
    \includegraphics[width=\figwidth]{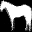} &
    \includegraphics[width=\figwidth]{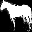}\\
    \includegraphics[width=\figwidth]{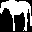} &
    \includegraphics[width=\figwidth]{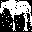} &
    \includegraphics[width=\figwidth]{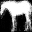} &
    \includegraphics[width=\figwidth]{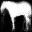} &
    \includegraphics[width=\figwidth]{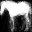} &
    \includegraphics[width=\figwidth]{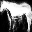} &
    \includegraphics[width=\figwidth]{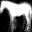} &
    \includegraphics[width=\figwidth]{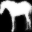} &
    \includegraphics[width=\figwidth]{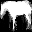} &
    \includegraphics[width=\figwidth]{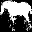}\\
    \includegraphics[width=\figwidth]{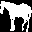} &
    \includegraphics[width=\figwidth]{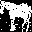} &
    \includegraphics[width=\figwidth]{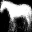} &
    \includegraphics[width=\figwidth]{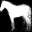} &
    \includegraphics[width=\figwidth]{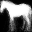} &
    \includegraphics[width=\figwidth]{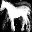} &
    \includegraphics[width=\figwidth]{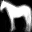} &
    \includegraphics[width=\figwidth]{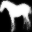} &
    \includegraphics[width=\figwidth]{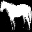} &
    \includegraphics[width=\figwidth]{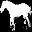}\\
    \includegraphics[width=\figwidth]{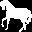} &
    \includegraphics[width=\figwidth]{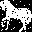} &
    \includegraphics[width=\figwidth]{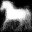} &
    \includegraphics[width=\figwidth]{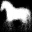} &
    \includegraphics[width=\figwidth]{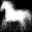} &
    \includegraphics[width=\figwidth]{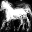} &
    \includegraphics[width=\figwidth]{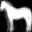} &
    \includegraphics[width=\figwidth]{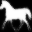} &
    \includegraphics[width=\figwidth]{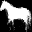} &
    \includegraphics[width=\figwidth]{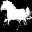}\\
    \includegraphics[width=\figwidth]{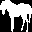} &
    \includegraphics[width=\figwidth]{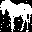} &
    \includegraphics[width=\figwidth]{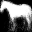} &
    \includegraphics[width=\figwidth]{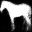} &
    \includegraphics[width=\figwidth]{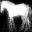} &
    \includegraphics[width=\figwidth]{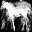} &
    \includegraphics[width=\figwidth]{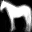} &
    \includegraphics[width=\figwidth]{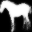} &
    \includegraphics[width=\figwidth]{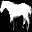} &
    \includegraphics[width=\figwidth]{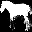}
    \end{tabular}
    \end{minipage}%
    \label{fig:generalisation_horses_10_noise}
    }\hfill%
    \subfigure[SSIM score (higher is better).]{%
\begin{minipage}[c]{0.26\linewidth}
\centering
\scalebox{0.8}{\small%
\begin{tabular}{lc}
Method & SSIM \\ \hline
DBN & $0.47 \pm 0.10$\\
SBM & $0.55 \pm 0.10$\\
VAE & $0.42 \pm 0.09$\\
InfoGAN & $0.33 \pm 0.10$\\
GPLVM & $0.44 \pm 0.07$\\
GPLVMDT & $0.54 \pm 0.09$\\[5pt]
\textbf{GPDBN} & $0.56 \pm 0.10$\\
\textbf{GPSBM} & $0.58 \pm 0.09$\\
\end{tabular}}
\label{tab:results_horses_10_noise}
%\end{table*}
\end{minipage}
    }\\[-10pt]%
    \caption{Manifold projection from corrupted observations. \subref{fig:generalisation_horses_10_noise}~Test silhouettes (first column) are corrupted with 10\% salt and pepper noise (second column). The remaining columns show estimated silhouettes from each model. \subref{tab:results_horses_10_noise}~Mean and standard deviation of the SSIM score between silhouettes from each model against the original test data without noise.}
\label{fig:generalisation_horses_10_noise_and_table}
\end{figure}

In addition to the results in the paper we report in Tab.~\ref{tab:results_horses_10_noise} the SSIM reconstruction score for 10\% salt-and-pepper noisy horses to demonstrates that our models consistently outperform the competition even with very little noise.

\begin{figure}[H]
\centering
\subfigure[Example results for projection onto manifold]{%
\begin{minipage}[c]{0.7\linewidth}
    \tiny\centering
    \setlength{\figwidth}{0.27in}
    \begin{tabular}{cccccccccc}
    \vspace{3pt}
    %Unseen & 20\% Noise & DBN & SBM & VAE & InfoGAN & GPLVM & GPLVMDT & GPDBN & GPSBM \\
    unseen & 20\% & dbn & sbm & vae & infogan & gpvlm & gplvmdt & gpdbn & gpsbm \\
    \includegraphics[width=\figwidth]{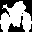} &
    \includegraphics[width=\figwidth]{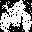} &
    \includegraphics[width=\figwidth]{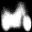} &
    \includegraphics[width=\figwidth]{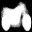} &
    \includegraphics[width=\figwidth]{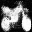} &
    \includegraphics[width=\figwidth]{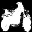} &
    \includegraphics[width=\figwidth]{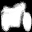} &
    \includegraphics[width=\figwidth]{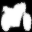} &
    \includegraphics[width=\figwidth]{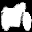} &
    \includegraphics[width=\figwidth]{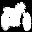}\\
    \includegraphics[width=\figwidth]{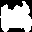} &
    \includegraphics[width=\figwidth]{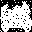} &
    \includegraphics[width=\figwidth]{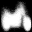} &
    \includegraphics[width=\figwidth]{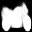} &
    \includegraphics[width=\figwidth]{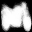} &
    \includegraphics[width=\figwidth]{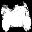} &
    \includegraphics[width=\figwidth]{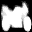} &
    \includegraphics[width=\figwidth]{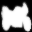} &
    \includegraphics[width=\figwidth]{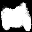} &
    \includegraphics[width=\figwidth]{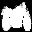}\\
    \includegraphics[width=\figwidth]{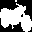} &
    \includegraphics[width=\figwidth]{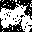} &
    \includegraphics[width=\figwidth]{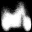} &
    \includegraphics[width=\figwidth]{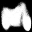} &
    \includegraphics[width=\figwidth]{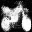} &
    \includegraphics[width=\figwidth]{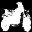} &
    \includegraphics[width=\figwidth]{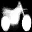} &
    \includegraphics[width=\figwidth]{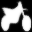} &
    \includegraphics[width=\figwidth]{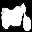} &
    \includegraphics[width=\figwidth]{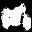}\\
    \includegraphics[width=\figwidth]{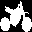} &
    \includegraphics[width=\figwidth]{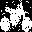} &
    \includegraphics[width=\figwidth]{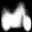} &
    \includegraphics[width=\figwidth]{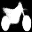} &
    \includegraphics[width=\figwidth]{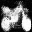} &
    \includegraphics[width=\figwidth]{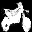} &
    \includegraphics[width=\figwidth]{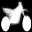} &
    \includegraphics[width=\figwidth]{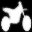} &
    \includegraphics[width=\figwidth]{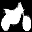} &
    \includegraphics[width=\figwidth]{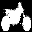}\\
    \includegraphics[width=\figwidth]{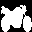} &
    \includegraphics[width=\figwidth]{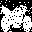} &
    \includegraphics[width=\figwidth]{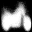} &
    \includegraphics[width=\figwidth]{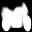} &
    \includegraphics[width=\figwidth]{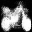} &
    \includegraphics[width=\figwidth]{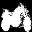} &
    \includegraphics[width=\figwidth]{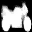} &
    \includegraphics[width=\figwidth]{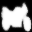} &
    \includegraphics[width=\figwidth]{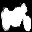} &
    \includegraphics[width=\figwidth]{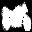}\\
    \includegraphics[width=\figwidth]{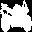} &
    \includegraphics[width=\figwidth]{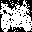} &
    \includegraphics[width=\figwidth]{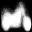} &
    \includegraphics[width=\figwidth]{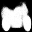} &
    \includegraphics[width=\figwidth]{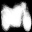} &
    \includegraphics[width=\figwidth]{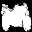} &
    \includegraphics[width=\figwidth]{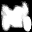} &
    \includegraphics[width=\figwidth]{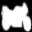} &
    \includegraphics[width=\figwidth]{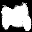} &
    \includegraphics[width=\figwidth]{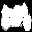}
    \end{tabular}
    \end{minipage}%
    \label{fig:generalisation_bikes_20_noise}
    }\hfill%
    \subfigure[SSIM score (higher is better).]{%
\begin{minipage}[c]{0.28\linewidth}
\centering
\scalebox{0.8}{\small%
\begin{tabular}{lc}
Method & SSIM \\ \hline
DBN & $0.42 \pm 0.03$\\
SBM & $0.58 \pm 0.04$\\
VAE & $0.35 \pm 0.11$\\
InfoGAN & $0.54 \pm 0.05$\\
GPLVM & $0.57 \pm 0.04$\\
GPLVMDT & $0.58 \pm 0.04$\\[5pt]
\textbf{GPDBN} & $0.64 \pm 0.03$\\
\textbf{GPSBN} & $0.63 \pm 0.02$\\
\end{tabular}}
%\caption{}
\label{tab:results_bikes_20_noise}
%\end{table*}
\end{minipage}
    }\\[-10pt]%
    \caption{Manifold projection from corrupted observations. \subref{fig:generalisation_bikes_20_noise}~Test silhouettes (first column) are corrupted with 20\% salt and pepper noise (second column). The remaining columns show estimated silhouettes from each model. \subref{tab:results_bikes_20_noise}~Mean and standard deviation of the SSIM score between silhouettes from each model against the original test data without noise. }
\label{fig:generalisation_bikes_20_noise_and_table}
\end{figure}

\begin{figure}[H]
\centering
\subfigure[Example results for projection onto manifold]{%
\begin{minipage}[c]{0.7\linewidth}
    \tiny\centering
    \setlength{\figwidth}{0.27in}
    \begin{tabular}{cccccccccc}
    \vspace{3pt}
    %Unseen & 40\% Noise & DBN & SBM & VAE & InfoGAN & GPLVM & GPLVMDT & GPDBN & GPSBM \\
    unseen & 40\% & dbn & sbm & vae & infogan & gpvlm & gplvmdt & gpdbn & gpsbm \\
    \includegraphics[width=\figwidth]{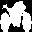} &
    \includegraphics[width=\figwidth]{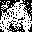} &
    \includegraphics[width=\figwidth]{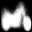} &
    \includegraphics[width=\figwidth]{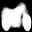} &
    \includegraphics[width=\figwidth]{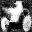} &
    \includegraphics[width=\figwidth]{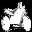} &
    \includegraphics[width=\figwidth]{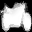} &
    \includegraphics[width=\figwidth]{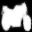} &
    \includegraphics[width=\figwidth]{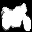} &
    \includegraphics[width=\figwidth]{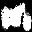}\\
    \includegraphics[width=\figwidth]{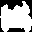} &
    \includegraphics[width=\figwidth]{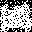} &
    \includegraphics[width=\figwidth]{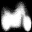} &
    \includegraphics[width=\figwidth]{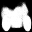} &
    \includegraphics[width=\figwidth]{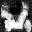} &
    \includegraphics[width=\figwidth]{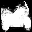} &
    \includegraphics[width=\figwidth]{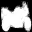} &
    \includegraphics[width=\figwidth]{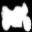} &
    \includegraphics[width=\figwidth]{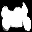} &
    \includegraphics[width=\figwidth]{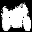}\\
    \includegraphics[width=\figwidth]{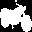} &
    \includegraphics[width=\figwidth]{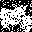} &
    \includegraphics[width=\figwidth]{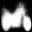} &
    \includegraphics[width=\figwidth]{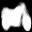} &
    \includegraphics[width=\figwidth]{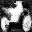} &
    \includegraphics[width=\figwidth]{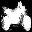} &
    \includegraphics[width=\figwidth]{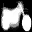} &
    \includegraphics[width=\figwidth]{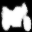} &
    \includegraphics[width=\figwidth]{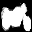} &
    \includegraphics[width=\figwidth]{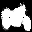}\\
    \includegraphics[width=\figwidth]{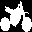} &
    \includegraphics[width=\figwidth]{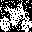} &
    \includegraphics[width=\figwidth]{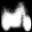} &
    \includegraphics[width=\figwidth]{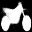} &
    \includegraphics[width=\figwidth]{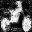} &
    \includegraphics[width=\figwidth]{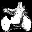} &
    \includegraphics[width=\figwidth]{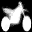} &
    \includegraphics[width=\figwidth]{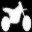} &
    \includegraphics[width=\figwidth]{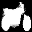} &
    \includegraphics[width=\figwidth]{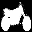}\\
    \includegraphics[width=\figwidth]{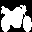} &
    \includegraphics[width=\figwidth]{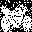} &
    \includegraphics[width=\figwidth]{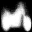} &
    \includegraphics[width=\figwidth]{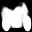} &
    \includegraphics[width=\figwidth]{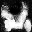} &
    \includegraphics[width=\figwidth]{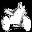} &
    \includegraphics[width=\figwidth]{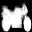} &
    \includegraphics[width=\figwidth]{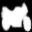} &
    \includegraphics[width=\figwidth]{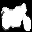} &
    \includegraphics[width=\figwidth]{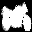}\\
    \includegraphics[width=\figwidth]{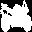} &
    \includegraphics[width=\figwidth]{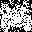} &
    \includegraphics[width=\figwidth]{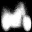} &
    \includegraphics[width=\figwidth]{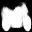} &
    \includegraphics[width=\figwidth]{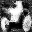} &
    \includegraphics[width=\figwidth]{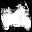} &
    \includegraphics[width=\figwidth]{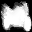} &
    \includegraphics[width=\figwidth]{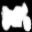} &
    \includegraphics[width=\figwidth]{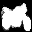} &
    \includegraphics[width=\figwidth]{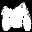}
    \end{tabular}
    \end{minipage}%
    \label{fig:generalisation_bikes_40_noise}
    }\hfill%
    \subfigure[SSIM score (higher is better).]{%
\begin{minipage}[c]{0.28\linewidth}
\centering
\scalebox{0.8}{\small%
\begin{tabular}{lc}
Method & SSIM \\ \hline
DBN & $0.42 \pm 0.03$\\
SBM & $0.57 \pm 0.04$\\
VAE & $0.17 \pm 0.02$\\
InfoGAN & $0.45 \pm 0.04$\\
GPLVM & $0.55 \pm 0.04$\\
GPLVMDT & $0.55 \pm 0.04$\\[5pt]
\textbf{GPDBN} & $0.58 \pm 0.05$\\
\textbf{GPSBN} & $0.61 \pm 0.03$\\
\end{tabular}}
%\caption{}
\label{tab:results_bikes_40_noise}
%\end{table*}
\end{minipage}
    }\\[-10pt]%
    \caption{Manifold projection from corrupted observations. \subref{fig:generalisation_bikes_40_noise}~Test silhouettes (first column) are corrupted with 40\% salt and pepper noise (second column). The remaining columns show estimated silhouettes from each model. \subref{tab:results_bikes_40_noise}~Mean and standard deviation of the SSIM score between silhouettes from each model against the original test data without noise. }
\label{fig:generalisation_bikes_40_noise_and_table}
\end{figure}

\begin{figure}[t]
\centering
\subfigure[Example results for projection onto manifold]{%
\begin{minipage}[c]{0.7\linewidth}
    \tiny\centering
    \setlength{\figwidth}{0.27in}
    \begin{tabular}{cccccccccc}
    \vspace{3pt}
    %Unseen & 60\% Noise & DBN & SBM & VAE & InfoGAN & GPLVM & GPLVMDT & GPDBN & GPSBM \\
    unseen & 60\% & dbn & sbm & vae & infogan & gpvlm & gplvmdt & gpdbn & gpsbm \\
    \includegraphics[width=\figwidth]{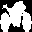} &
    \includegraphics[width=\figwidth]{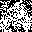} &
    \includegraphics[width=\figwidth]{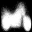} &
    \includegraphics[width=\figwidth]{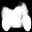} &
    \includegraphics[width=\figwidth]{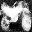} &
    \includegraphics[width=\figwidth]{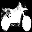} &
    \includegraphics[width=\figwidth]{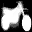} &
    \includegraphics[width=\figwidth]{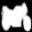} &
    \includegraphics[width=\figwidth]{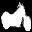} &
    \includegraphics[width=\figwidth]{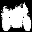}\\
    \includegraphics[width=\figwidth]{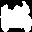} &
    \includegraphics[width=\figwidth]{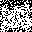} &
    \includegraphics[width=\figwidth]{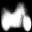} &
    \includegraphics[width=\figwidth]{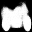} &
    \includegraphics[width=\figwidth]{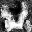} &
    \includegraphics[width=\figwidth]{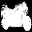} &
    \includegraphics[width=\figwidth]{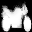} &
    \includegraphics[width=\figwidth]{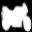} &
    \includegraphics[width=\figwidth]{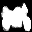} &
    \includegraphics[width=\figwidth]{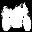}\\
    \includegraphics[width=\figwidth]{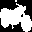} &
    \includegraphics[width=\figwidth]{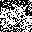} &
    \includegraphics[width=\figwidth]{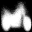} &
    \includegraphics[width=\figwidth]{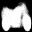} &
    \includegraphics[width=\figwidth]{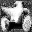} &
    \includegraphics[width=\figwidth]{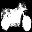} &
    \includegraphics[width=\figwidth]{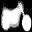} &
    \includegraphics[width=\figwidth]{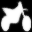} &
    \includegraphics[width=\figwidth]{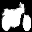} &
    \includegraphics[width=\figwidth]{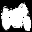}\\
    \includegraphics[width=\figwidth]{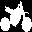} &
    \includegraphics[width=\figwidth]{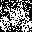} &
    \includegraphics[width=\figwidth]{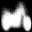} &
    \includegraphics[width=\figwidth]{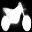} &
    \includegraphics[width=\figwidth]{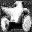} &
    \includegraphics[width=\figwidth]{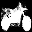} &
    \includegraphics[width=\figwidth]{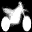} &
    \includegraphics[width=\figwidth]{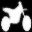} &
    \includegraphics[width=\figwidth]{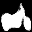} &
    \includegraphics[width=\figwidth]{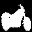}\\
    \includegraphics[width=\figwidth]{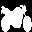} &
    \includegraphics[width=\figwidth]{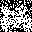} &
    \includegraphics[width=\figwidth]{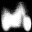} &
    \includegraphics[width=\figwidth]{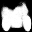} &
    \includegraphics[width=\figwidth]{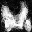} &
    \includegraphics[width=\figwidth]{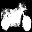} &
    \includegraphics[width=\figwidth]{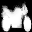} &
    \includegraphics[width=\figwidth]{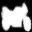} &
    \includegraphics[width=\figwidth]{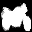} &
    \includegraphics[width=\figwidth]{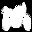}\\
    \includegraphics[width=\figwidth]{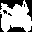} &
    \includegraphics[width=\figwidth]{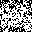} &
    \includegraphics[width=\figwidth]{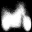} &
    \includegraphics[width=\figwidth]{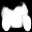} &
    \includegraphics[width=\figwidth]{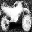} &
    \includegraphics[width=\figwidth]{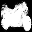} &
    \includegraphics[width=\figwidth]{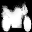} &
    \includegraphics[width=\figwidth]{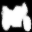} &
    \includegraphics[width=\figwidth]{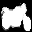} &
    \includegraphics[width=\figwidth]{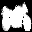}
    \end{tabular}
    \end{minipage}%
    \label{fig:generalisation_bikes_60_noise}
    }\hfill%
    \subfigure[SSIM score (higher is better).]{%
\begin{minipage}[c]{0.28\linewidth}
\centering
\scalebox{0.8}{\small%
\begin{tabular}{lc}
Method & SSIM \\ \hline
DBN & $0.42 \pm 0.04$\\
SBM & $0.55 \pm 0.05$\\
VAE & $0.17 \pm 0.03$\\
InfoGAN & $0.50 \pm 0.04$\\
GPLVM & $0.51 \pm 0.06$\\
GPLVMDT & $0.54 \pm 0.05$\\[5pt]
\textbf{GPDBN} & $0.55 \pm 0.07$\\
\textbf{GPSBN} & $0.55 \pm 0.07$\\
\end{tabular}}
%\caption{}
\label{tab:results_bikes_60_noise}
%\end{table*}
\end{minipage}
    }\\[-10pt]%
    \caption{Manifold projection from corrupted observations. \subref{fig:generalisation_bikes_60_noise}~Test silhouettes (first column) are corrupted with 60\% salt and pepper noise (second column). The remaining columns show estimated silhouettes from each model. \subref{tab:results_bikes_60_noise}~Mean and standard deviation of the SSIM score between silhouettes from each model against the original test data without noise. }
\label{fig:generalisation_bikes_60_noise_and_table}
\end{figure}

Fig.~\ref{fig:generalisation_bikes_20_noise_and_table}, Fig.~\ref{fig:generalisation_bikes_40_noise_and_table} and Fig.~\ref{fig:generalisation_bikes_60_noise_and_table} show the results of an experiment where test data has been corrupted by significant noise (20\%, 40\% and 60\% respectively) and we wish to project onto the manifold of valid silhouettes. The quantitative comparisons indicate that our models have managed to capture a good probabilistic estimate of the data manifold while still preserving interpretability.

\begin{figure}[H]
\centering
\includegraphics[width=0.85\linewidth]{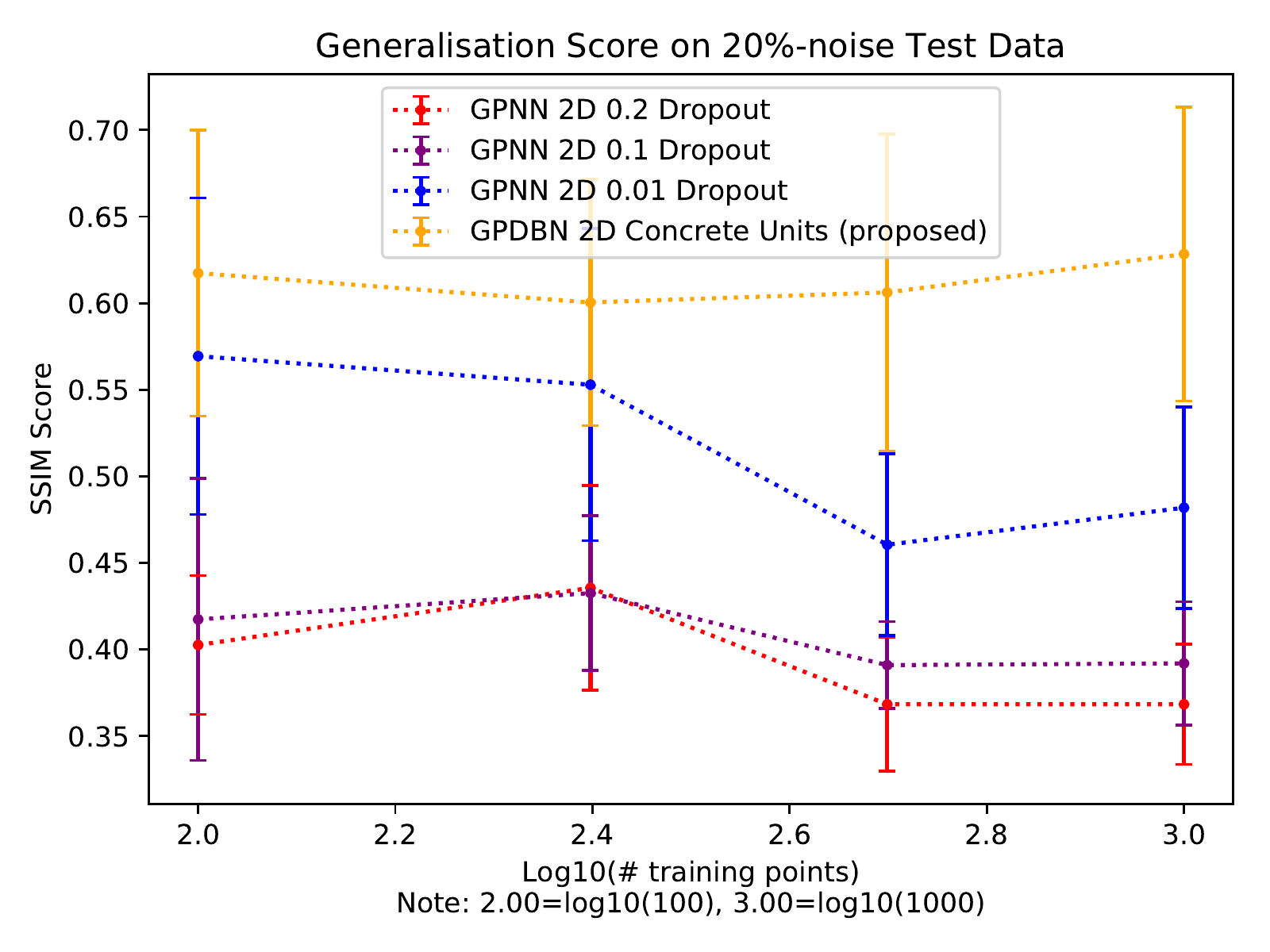}\\[-10pt]
\caption{figure}{Ablation comparisons of GPDBN models variants using sigmoidal units plus Dropout, showing the importance of Concrete units for proper uncertainty propagation. (A \emph{GPNN} architecture is exactly like a GPDBN except that it uses sigmoidal units.)}
\label{fig:gpdbn_fixed_dropout}
\end{figure}

\begin{table}[H]
\centering
\begin{tabular}{lc}
Method & SSIM \\ \hline
Large net (x3 units) & $0.52 \pm 0.12$ \\
Narrow net (1/3 units) & $0.47 \pm 0.15$ \\
Deep net (+1 layer) & $0.45 \pm 0.20$ \\
Shallow net (-1 layer) & $0.58 \pm 0.08$\\
\end{tabular}
\caption{SSIM of the output (without 20\% salt-and-pepper noise) of the GPDBN using different network architectures.}
\label{tab:different_architectures}
\end{table}

\subsection{Robustness and Ablation}
Given the space constraints in the paper we have given priority to what we believe is the most important; we have provided extensive results and comparisons with recents models (plus an interactive demo). In the following paragraphs we address feature ablation and model robustness.

We note that the two components of our GPDBN model, that is the
GPLVM and DBN, are both essential, none of them can be ablated because the former provides the smooth manifold and
predictive uncertainty while the latter increases the capability of generating image data.
Ablation of our important Concrete units to dropout units impedes uncertainty propagation and reduces performance (Fig.~\ref{fig:gpdbn_fixed_dropout}).
Moreover, our comparisons show that both GPLVM and DBN are weaker as standalone models.

The architectures used were based on previous work to enable fair comparison.
In addition, we have trained a GPDBN on the horse dataset experimenting with four different networks (increasing and decreasing the number of units and layers).
By removing one layer (Tab.~\ref{tab:different_architectures}) we got slightly better performance to the more generic architecture proposed in the paper.
It might be that a shallow network is better suited for this data given the small number of training examples.
We see this positively as the model can be fine-tuned to achieve even higher performance depending on the specific data.
Finding the optimal number of layers and weights (parameters) is an open issue common to many deep learning methods.
We think that replacing the GPLVM part with a Bayesian one (\cite{titsias2010bayesian}) would solve this problem for the GPDBN allowing it to use the optimal number of parameters automatically. %This is a good suggestion for future investigation.

\subsection{DBNs and Mini-batching} Traditionally, DBNs do not train on large images, this is because of the high number of parameters %(a lot of memory power would be required).
therefore, the use of convolutions is necessary.
In contrast, GPDBN mini-batching allows us to deal with a large number of images and empirically any introduced bias does not really reduce the performance. For example, we trained a GPDBN on $1000$ MNIST digits (as in Fig 6 in the paper) with mini-batch size $100$, we obtained SSIM: $0.63 \pm 0.09$, which is even higher than the non mini-batched equivalent (SSIM: $0.50 \pm 0.08$).

\subsection{Smoothness Experiment}

\begin{figure*}[t!]
\centering
\setlength{\tabcolsep}{2pt}
    \includegraphics[width=0.9\linewidth]{figures/star-dataset-border.png}\\[4pt]
    \begin{tabular}{cccc}
    GPDBN & GPSBM & GPLVM & GPLVMDT \\
    \includegraphics[width=0.23\linewidth]{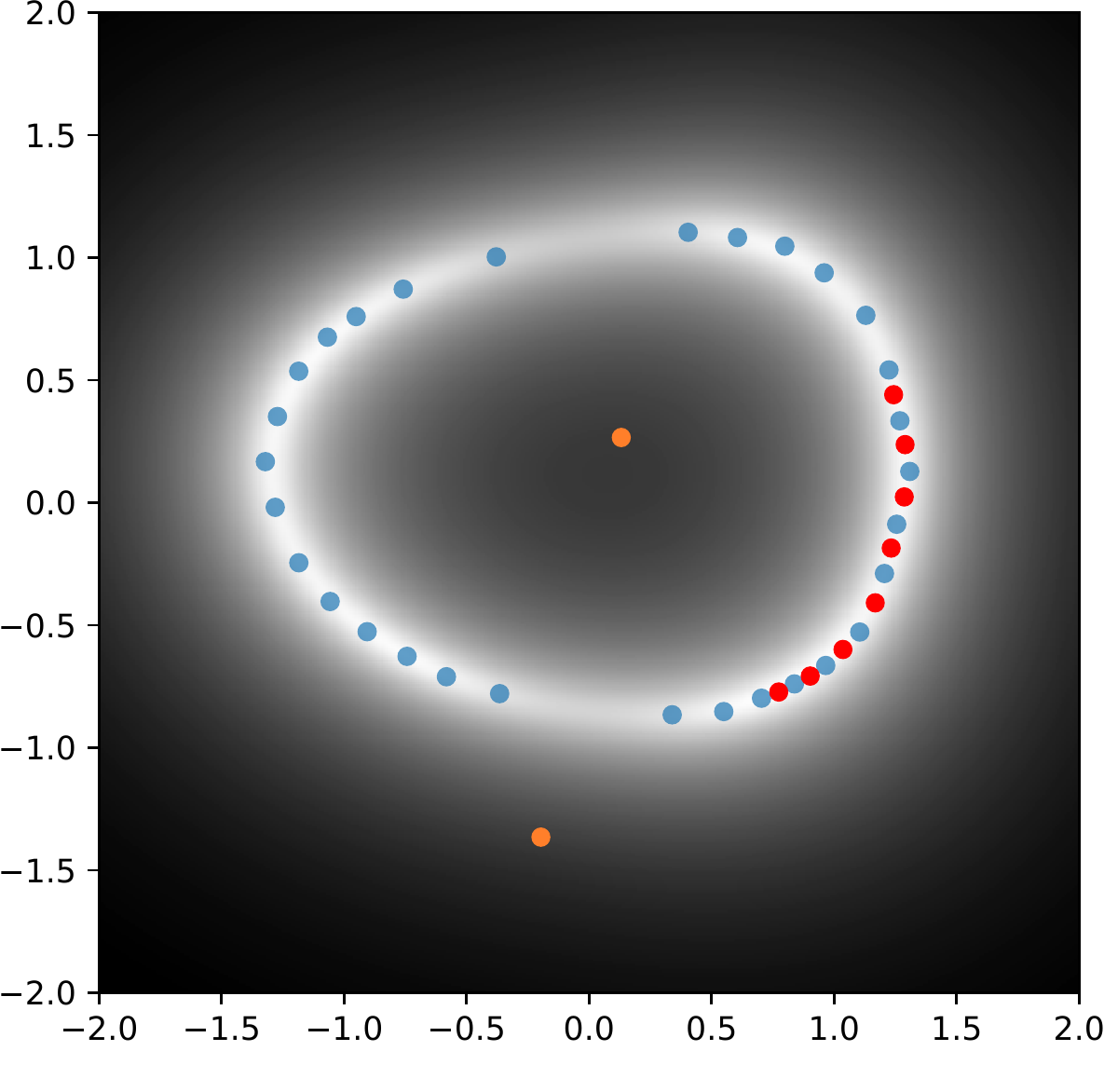} &
    \includegraphics[width=0.23\linewidth]{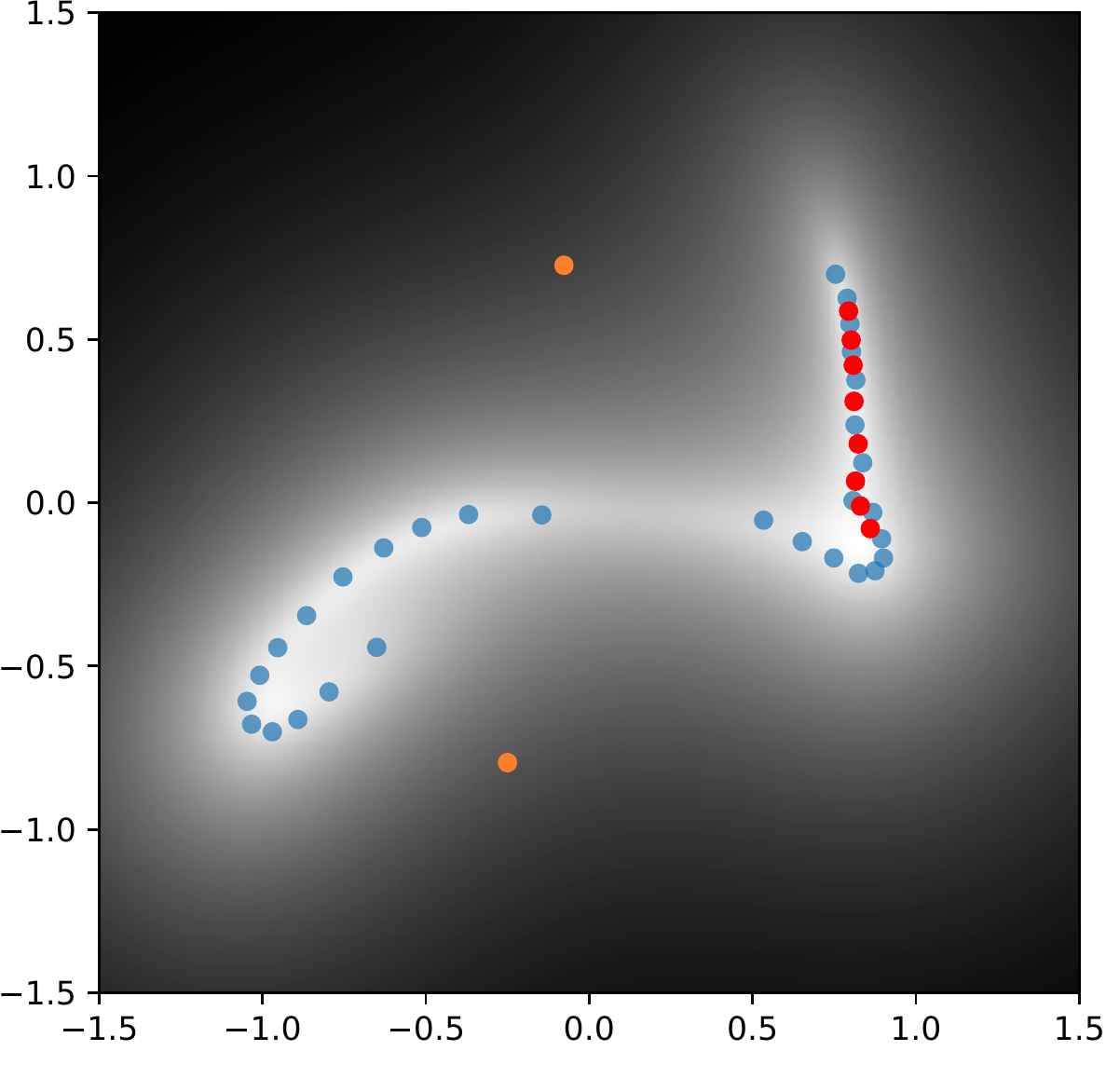} &
    \includegraphics[width=0.23\linewidth]{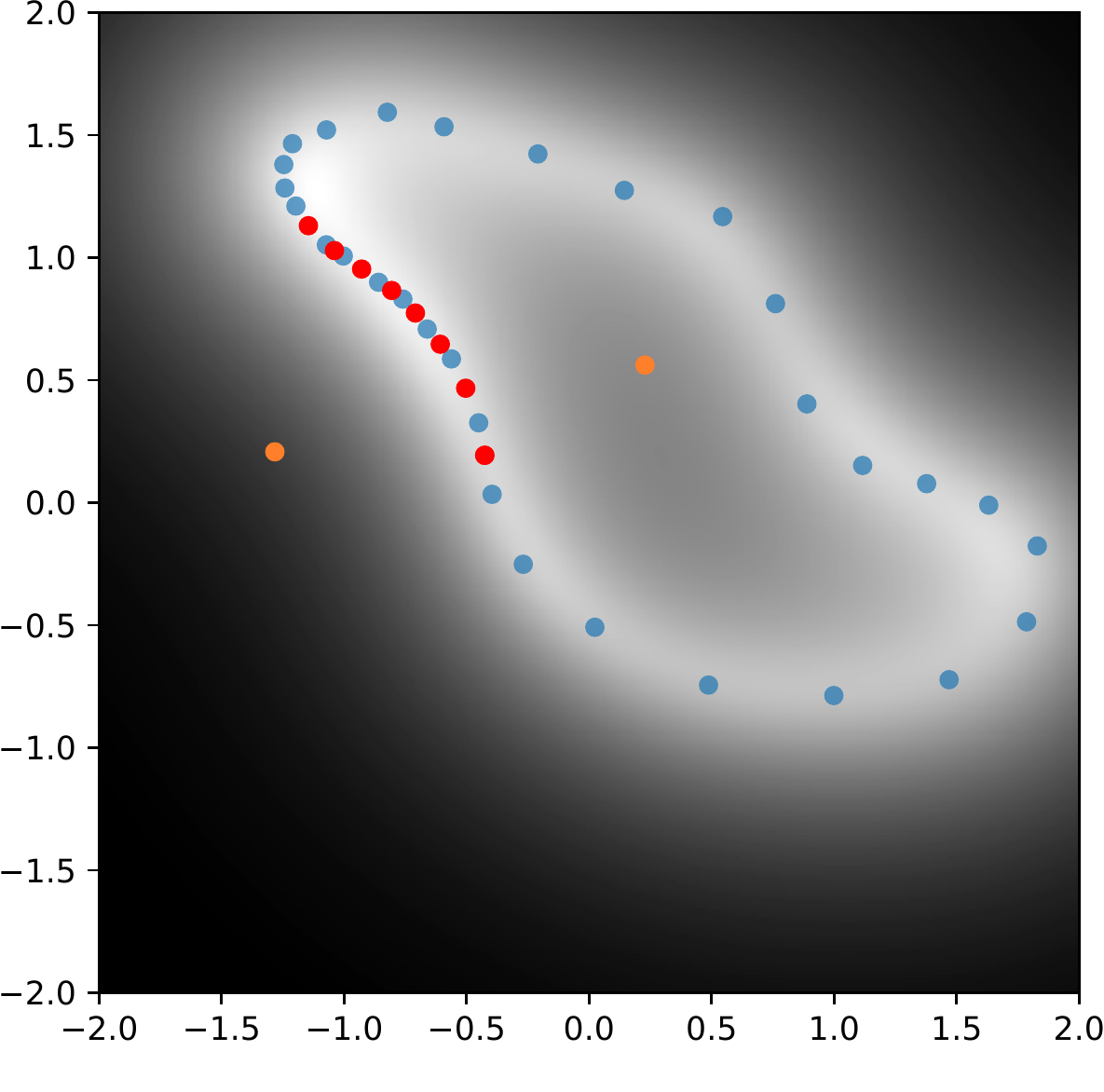} &
    \includegraphics[width=0.23\linewidth]{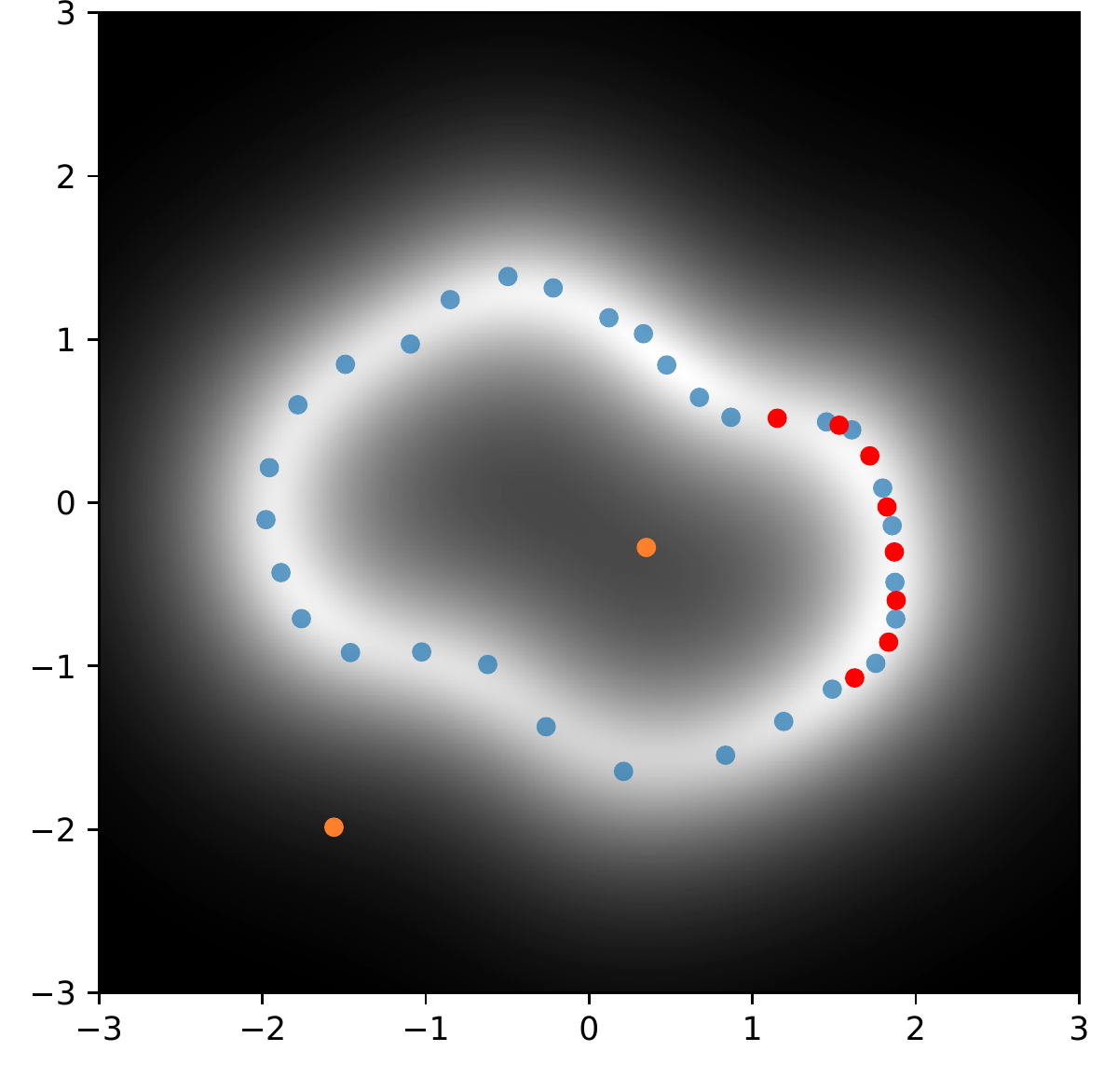} \\
    %\rule{0pt}{5ex}
    \includegraphics[width=0.24\linewidth]{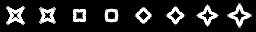} &
    \includegraphics[width=0.24\linewidth]{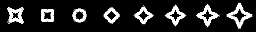} &
    \includegraphics[width=0.24\linewidth]{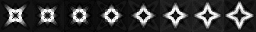} &
    \includegraphics[width=0.24\linewidth]{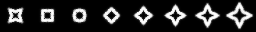} \\
    \rule{0pt}{6ex}
    \includegraphics[width=0.1\linewidth]{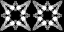} &
    \includegraphics[width=0.1\linewidth]{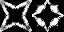} &
    \includegraphics[width=0.1\linewidth]{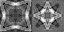} &
    \includegraphics[width=0.1\linewidth]{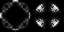}
    \end{tabular}\\[-5pt]
\caption{Comparison of manifold models trained on the star dataset that consists of $30$ points on a 1D manifold mapped to a corresponding set of silhouettes (top row).
 The \textcolor{blue}{blue} points on each manifold are the latent locations corresponding to the training examples from the top row. The $8$ \textcolor{red}{red} points are test locations on a smooth path on the manifold; the corresponding $8$ novel silhouettes generated at these points are shown below each manifold. The bottom row shows $2$ silhouettes corresponding to the $2$ \textcolor{orange}{orange} points outside the manifold.}
\label{fig:manifolds}
\end{figure*}

This additional experiment also highlights the benefit of the uncertainty associated with our model and how it manifests itself in the proposed GPDBN in contrast with the other methods. There is an inherent trade-off between a simple topology of the manifold and the smoothness of the mapping. To exemplify this we generated a dataset of a shape deformed in a cyclic manner Fig.~\ref{fig:manifolds}. The resulting latent space is structured as a circle clearly reflecting the topology of the deformation. Importantly, if the uncertainty in the model reflects that of the data we should move along ridges of high probability (manifold geodesics) to generate realistic data. Our proposed model is directly applicable to such approaches as described in~\cite{Tosi:2014tt}. Further, the experiment highlights how the uncertainty effects the prediction. When generating shapes corresponding to a region of the manifold where the model is highly uncertain we would, if the model have captured the characteristics of the data well,  expect images corresponding to the \emph{average} shape. As can be seen the GPDBN clearly generates the average shape while the other methods fail to capture this characteristic in the data making it challenging to interpret the uncertainty. %In Fig.~\ref{fig:scalability} we show results on the MNIST data-set~\cite{lecun-mnisthandwrittendigit-2010} as can be seen when ignoring the uncertainty transformations between digits will result in shapes not corresponding to digits. By making use of the uncertainty such cases can be avoided.

%% We created a toy dataset to demonstrate that the GPDBN retains the smooth manifold property of the GPLVM by influencing the learned representation from the DBN. Fig.~\ref{fig:manifolds} shows the dataset in the top row; we generated $30$ silhouettes from a smooth 1D manifold that parameterises the points of an octohedron. Ideally, we would like a model to recover this smooth manifold by encoding the training data on a continuous 1D trajectory that interpolates the training examples smoothly and correctly.
%% We note again that, while the dataset appears smooth to us, in pixel space the transitions are far from smooth; the Euclidean distance, or a Gaussian likelihood, in this vector space does not produce a smooth mapping.

\paragraph{Structure}
The blue points on the manifolds in Fig.~\ref{fig:manifolds} show the results for each of the smooth manifold based models. We note that all the manifolds have correctly identified a smooth trajectory for the training data. In addition, all but the GPSBM have captured the periodic repetition by closing the path; this is possibly due to the symmetry in the dataset not reflecting the shared architecture of the SBM.

The red points represent test locations corresponding to the samples in the third row of Fig.~\ref{fig:manifolds}.
Here we see that all models are correctly interpolating the overall pattern, however, the Gaussian likelihood of the GPLVM introduces artefacts in the silhouettes that are not found in the results from the GPDBN and GPSBM. The GPLVMDT improves over the GPLVM but still produces blurred results.

\paragraph{Uncertainty}
Finally, the real power of the GPDBN model is captured by looking at what happens when you leave the manifold. The final row of silhouettes are samples from the orange points that are in regions of high predictive variance (low confidence). Both the GPLVM and the GPLVMDT produce completely unreasonable results. Whereas the GPDBN \emph{captures the uncertainty in the manifold perfectly}; we see the average probability of the entire dataset with the predictive probability correctly captured. The GPDBN results are the mean of a set of samples from the model and away from the manifold these results are correctly approaching the mean of the training data.
Interestingly, the results for the GPSBM show the asymmetry in the shared weights; leaving the manifold in two different directions averages different regions of the training data.

\end{document}